\documentclass[10pt,journal,twoside,letterpaper,compsoc]{IEEEtran}

\usepackage[nocompress]{cite}

\usepackage[pdftex]{graphicx}
\usepackage{epstopdf}
\usepackage{microtype}

\usepackage{amssymb}
\usepackage[cmex10]{amsmath}
\usepackage{array}
\usepackage{mdwmath}
\usepackage{mdwtab}
\usepackage{eqparbox}
\usepackage{multirow}

\usepackage[caption=false,font=footnotesize,labelfont=sf,textfont=sf]{subfig}
\usepackage{stfloats}
\usepackage{url}

\usepackage{rotating}
\usepackage{setspace}
\usepackage{pifont}

\begin{document}

\title{Box-Free Model Watermarks Are Prone to Black-Box Removal Attacks}

\author{Haonan An,~\IEEEmembership{Student Member,~IEEE,} 
        Guang Hua,~\IEEEmembership{Senior Member,~IEEE,}\\
        Zhiping Lin,~\IEEEmembership{Senior Member,~IEEE,}
        and Yuguang Fang,~\IEEEmembership{Fellow,~IEEE}

\IEEEcompsocitemizethanks{
    \IEEEcompsocthanksitem Haonan An and Zhiping Lin are with the School of Electrical and Electronic Engineering, Nanyang Technological University, Singapore 639798.\protect\\
    E-mail: an0029an@e.ntu.edu.sg, ezplin@ntu.edu.sg.
    \IEEEcompsocthanksitem Guang Hua is with the Infocomm Technology (ICT) Cluster, Singapore Institute
of Technology (SIT), Singapore 138683. \protect\\
    E-mail: ghua@ieee.org, guang.hua@singaporetech.edu.sg.
    \IEEEcompsocthanksitem Yuguang Fang is with the Department of Computer Science, City University of Hong Kong, Hong Kong. \protect\\
    E-mail: my.fang@cityu.edu.hk. 
    }%
\thanks{The work was supported in part by the Hong Kong SAR Government under the Global STEM Professorship and the Hong Kong Jockey Club under JC STEM Lab of Smart City.   }
\thanks{(Corresponding Author: Guang Hua.)}

}

\markboth{}
{An \MakeLowercase{\textit{et al.}}: Box-Free Model Watermarks Are Prone to Black-Box Removal Attacks}


\IEEEtitleabstractindextext{
\begin{abstract}
Box-free model watermarking is an emerging technique to safeguard the intellectual property of deep learning models, particularly those for low-level image processing tasks. Existing works have verified and improved its effectiveness in several aspects. However, in this paper, we reveal that box-free model watermarking is prone to removal attacks, even under the real-world threat model such that the protected model and the watermark extractor are in black boxes. Under this setting, we carry out three studies. 1) We develop an extractor-gradient-guided (EGG) remover and show its effectiveness when the extractor uses ReLU activation only. 2) More generally, for an unknown extractor, we leverage adversarial attacks and design the EGG remover based on the estimated gradients. 3) Under the most stringent condition that the extractor is inaccessible, we design a transferable remover based on a set of private proxy models. In all cases, the proposed removers can successfully remove embedded watermarks while preserving the quality of the processed images, and we also demonstrate that the EGG remover can even replace the watermarks. Extensive experimental results verify the effectiveness and generalizability of the proposed attacks, revealing the vulnerabilities of the existing box-free methods and calling for further research.

\end{abstract}

}

\maketitle

\section{Introduction}
\IEEEPARstart{F}{o}llowing the success of AlexNet in 2012, deep neural networks (DNNs) have undergone rapid development, notably contributing in areas such as natural language processing \cite{devlin2018bert}, image recognition \cite{dosovitskiy2020image}, healthcare \cite{lam2018automated}, and other fields. DNNs operate on extensive datasets and require significant computing, communications, and energy resources to be effectively trained for specific tasks. For instance, current advanced generative models like ChatGPT-3, which boasts 175 billion parameters \cite{brown2020language}, involve a large amount of monetary investments in data collection and network training. As a result, these models are increasingly viewed as precious assets and constitute a novel form of intellectual property that warrants careful safeguarding.

Common goals of DNN intellectual property protection are ownership verification and model stealing (also known as model extraction or surrogate attack) tracking. Both goals can be effectively achieved through DNN watermarking \cite{li2021survey, zhong2023brief, sharma2023review}. Existing DNN watermarking methods can be classified into white-box, black-box, and box-free methods \cite{li2021survey}. In white-box methods, the watermark is embedded directly into the model's parameters, which requires access to the model's internal structure for watermark extraction. While this approach allows for accurate watermark extraction due to direct access to the model parameters, it sacrifices the model privacy. Meanwhile, providing access to a model's structure and parameters for extraction may also violate security protocols, leading to the potential alteration or removal of the watermark. In contrast, black-box methods circumvent the need to access the model's internal information. Instead, they employ specific inputs, referred to as triggers, which can alter the model's behavior. This is also termed as backdoor watermarking, and the verification is done by analyzing model inputs and outputs.

The existing white- and black-box methods are primarily employed for the protection of classification models. We note that classification models only generate low entropic outputs, e.g., the categorical confidence scores, or even merely the top-1 label. This aligns with the essence of white- and black-box methods that embed watermarks into model parameters and behaviors, respectively. In contrast, for generative models, e.g., image processing models we focus on in this paper, the outputs contain high entropic image content, allowing the models to generate watermarked outputs, known as box-free watermarking \cite{zhang2020model}. In this type of methods, a watermark hiding network and an extraction network are respectively designed and can be trained jointly with the original model. Then, the protected model (original model plus hiding network) generates processed and watermarked images, while the extraction network, when queried by (stolen) model-generated images, yields the watermark, or otherwise yields a noisy or non-informative pattern. Box-free methods excel at their flexibility as they do not require access to model parameters. The output images are also called containers \cite{zhang2020model}. 

The existing works on box-free methods have focused on watermark fidelity, robustness against surrogate attacks, robustness against normal image processing, and extension to different generative tasks \cite{wu2020watermarking,zhang2021deep,huang2023can}. Despite the current research progress, in this paper, we make the first attempt to reveal an overlooked vulnerability of box-free methods against black-box watermark removal attacks. We focus exclusively on deep generative models for low-level image processing tasks and consider three scenarios. \textbf{Scenario 1:} The victim model, consisting of an image processing network (IPNet), a watermark hiding network (HNet), and a watermark extraction network (ENet), is secured in a black box, whose API are available for query, and ENet only uses ReLU as activation. We then propose an extractor-gradient-guided (EGG) watermark removal attack and show its effectiveness. \textbf{Scenario 2:} It reflects a more challenging yet practical situation which is almost the same as Scenario 1 except that ENet uses arbitrary and unknown activation functions. In this scenario, ENet's gradients can be estimated via querying, according to studies in adversarial attacks\footnote{In black-box adversarial attack literature \cite{chen2017zoo, dong2021query, shi2022query}, the gradient information of the black-box victim model is estimated via querying. While this is beyond the scope of this paper, it validates our Scenario 2 in which the gradients of ENet can be reverse-engineered.}, and we develop an EGG remover under this condition.  \textbf{Scenario 3:} It is the most challenging scenario in which the victim model is in a black box while ENet API is not provided. Since the original functionality of IPNet is known, we resort to a transferability-based attack, which trains a private set of proxy models denoted by $\text{HNet}'$, and $\text{ENet}'$, respectively. Then, we show that the remover trained for the proxy models can be transferred to attack the victim model.

In the above three scenarios, the proposed EGG removers, as well as the transferable remover, when fed with container (processed and watermarked) images, can effectively remove the embedded watermarks while preserving the image quality. We note that the proposed removers are also image-to-image models, which turn a container image into a watermark-free one. This differs from existing image-processing-based removal (e.g., adding noise, performing transformations, etc., usually less effective) \cite{lukas2022sok} and white-box ones (e.g., fine-tuning and compression, less practical than black-box ones) \cite{guo2021fine}.
The contributions of this paper are summarized as follow:
\begin{itemize}
    \item We reveal the vulnerabilities of the existing box-free model watermarking methods against black-box removal attacks. It takes a distinctive perspective compared to most existing works dealing only with white-box removal such as fine-tuning and compression. This indicates that even black-box protection cannot effectively prevent watermark removal.
    \item We analyze the vulnerabilities by considering three scenarios in our threat model, reflecting different attack difficulties. Via the proposed EGG and transferable removers, removal attacks can be launched in all situations, reflecting real-world threats.
    \item We design and conduct extensive experiments to demonstrate that the proposed attacks can not only remove watermarks in container images, but also preserve image quality. This is achievable even under stringent attacking conditions. Moreover, we demonstrate that the EGG remover can even overwrite watermarks to mislead ownership verification.
\end{itemize}

\section{Related Work}
\subsection{White-Box Watermarking}
The white-box methods directly embed watermarks into model parameters (considered as the carrier in classic watermarking). It was first proposed by Uchida et al. \cite{uchida2017embedding}, where a regularization term is imposed during the learning of the original functionality to introduce a bias that allows the owner to extract the watermark from the model parameters using a preset projection matrix. After that, research efforts have been devoted to the exploration of embedding locations within the deep learning model, such as activation maps \cite{darvish2019deepsigns}, feature maps \cite{li2021feature}, decision boundary \cite{le2020adversarial}, and additional layers \cite{fan2021deepipr}. Notably, Li et al. \cite{li2021spread} incorporated the classic spread-transform dither-modulation for image watermarking in model watermarking. Overall, white-box methods are constrained by the need for the model's internal information for watermark extraction. 

\subsection{Black-Box Watermarking}
Black-box model watermarking is also known as backdoor watermarking, whose watermark extraction process only requires the observation of the trigger-output relationships. Most existing backdoor watermarks are in the form of trigger-label mappings designed specially for deep classification models. The triggers can be both in- and out-of-distribution samples \cite{darvish2019deepsigns, hua23unambiguous, adi2018turning, zhang2018protecting, namba2019exponential}, while the trigger labels can be drawn from either the original categories or a novel category \cite{zhong2020protecting, li2020protecting}. However, Huang et al. \cite{huang2023can} have pointed out the limitations of existing black-box methods in protecting generative models, noting that these methods are not applicable to noise-to-image models. In addition, black-box methods are unable to protect the generated images, which are also considered intellectual properties. In summary, while black-box watermarking could be more secure than white-box watermarking, it is unsuitable for safeguarding generative models. We note that Quan et al. \cite{quan2020watermarking} have made the first attempt to backdoor watermark generative models.

\subsection{Box-Free Watermarking}
Box-free watermarking was motivated by the need for protecting the high-entropic generated images \cite{li2021survey}. The term ``box-free'' originates from the minimal dependency of watermark extraction on the model itself.  In this line of research, Yu et al. \cite{yu2021artificial} proposed an indirect approach to protect image-to-image generative adversarial networks (GANs). It first watermarks the training data and then, based on transferability, extracts the watermarks from images generated by the trained GANs using a pretrained extractor. They further proposed a method to protect noise-to-image GANs, which embeds the watermark in the generator during training \cite{yu2022responsible}. Wu et al. \cite{wu2020watermarking} proposed a framework similar to \cite{yu2022responsible} but for image-to-image models, where watermark embedding and model functionality are jointly trained. Alternatively, Zhang et al. \cite{zhang2021deep} decoupled the two tasks by designing a dedicated hiding network that is encapsulated together with the original model. We note that the box-free watermarking considered in this paper differs from the subject of ``hiding images within images'' \cite{Baluja2020hiding}, as the former aims for model stealing tracking but the latter is for high-capacity data hiding. We will implement the state-of-the-art box-free methods \cite{wu2020watermarking} and \cite{zhang2021deep} for experiments.

Additionally, we note that the work in \cite{huang2023can} is a special case of box-free watermarking. It is a dedicated method for GANs leveraging the generator information left in the corresponding discriminator due to adversarial learning. Therefore, there is no explicit watermark embedding and extraction processes therein. Instead, ownership verification is realized by further training the discriminator as a classifier and querying it with a test image so that whether the image is generated by the protected generator can be detected. However, since the verification process is the same as those in the common box-free method \cite{wu2020watermarking}, \cite{zhang2021deep}, we will launch our proposed attacks against  \cite{huang2023can} in the experiments to test its generalization capability when the proxy framework is different from that in the victim model.  

\begin{figure}[!t]
  \centering
  \includegraphics[width=0.5\textwidth]{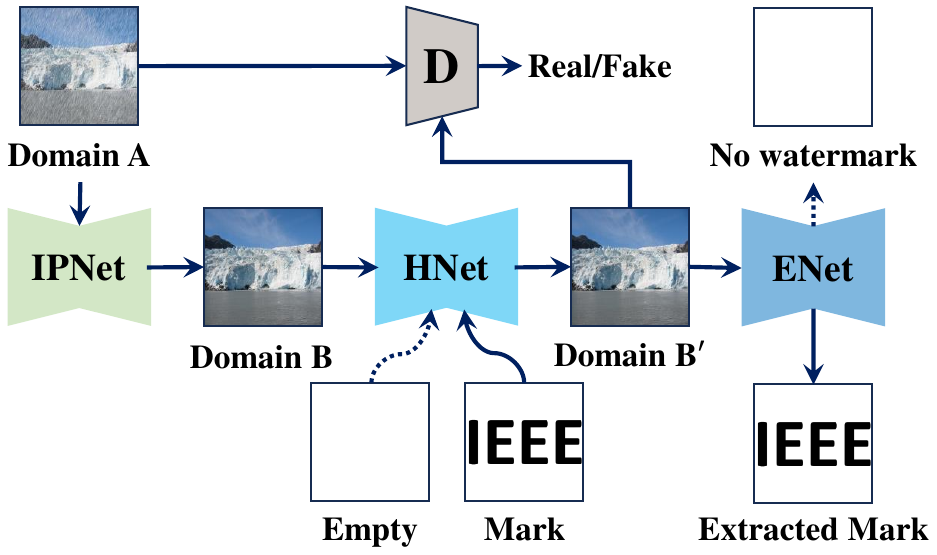}
  \vspace*{-6pt}
  \caption{Flowchart of box-free watermarking for image-to-image models, where image denoising is used as an example.}
  \label{fig:victim_model}
  \vspace{-3mm}
\end{figure}

\subsection{Watermark Removal}
Watermark removal attacks aim to remove watermarks embedded in model parameters, behaviors (backdoor), or outputs, and thus result in failure of extraction. It can be either a white-box or a black-box attack, with or without the victim model's internal information. Lukas et al. \cite{lukas2022sok} comprehensively investigated the robustness of model watermarks against removal attacks including image preprocessing, white-box model modification, and black-box model stealing. In this spirit, Wang et al. \cite{wang2022rethinking} incorporated adversarial perturbations for preprocessing-based removal, while Guo et al. \cite{guo2021fine} proposed an improved fine-tuning-based white-box removal attack. However, all these methods are only applicable to classification models and white-box setting.

For image processing models, removal is usually related to robustness, and existing works have considered watermark robustness against normal image preprocessing, e.g., cropping, blurring, compression, and adding noise \cite{yu2022responsible}, \cite{wu2020watermarking}. Recently, Liu et al. \cite{liu2023erase} developed a model agnostic lattice attack and an inpainting-based black-box remover. The lattice attack replaces image pixels at specific locations by random values to remove the watermark, but it results in visible artifacts. The inpainting-based attack, on the other hand, first removes a few image patches and then repairs them via deep inpainting. However, it requires the image before watermarking, i.e., the output of IPNet, to supervise the inpainter, which is not available in our black-box setting.


Other types of watermark removal attacks are noted here for completeness. Jiang et al. \cite{jiang2023evading} proposed the so-called WEvade-B-S, but with limited performance in their experiments, and WEvade-B-Q, a recursive attack needing multiple iterations to attack a single image, which is time and computational inefficient. You et al. \cite{you2024two} designed a dedicated remover to remove spread spectrum (SS) image watermarks, but it is a known message attack (KMA) requiring watermark information, which is not applicable in our black-box setting. Additionally, Niu et al. \cite{Niu2023fine}, Tian et al. \cite{Tian2024self, Tian2024Perceptive} studied the removal of visible watermarks, whereas our work targets the more challenging task of removing invisible watermarks. We also notice that the ambiguity attacks can confuse ownership verification by introducing multiple watermarks, but here we focus on the more formidable challenge of complete watermark removal.

\begin{table}[!t]
\renewcommand{\arraystretch}{1.0}
\centering
\caption{Summary of symbols and abbreviations.}
\label{tab:notation} 
\vspace*{-9pt}
\setlength{\tabcolsep}{4pt}
{\begin{tabular}{c|c}
\hline
\hline
Symbol/Abbr.  & Meaning \\
\hline
IPNet & Image processing network \\
HNet &  Hiding network\\
ONet &  Operation network (IPNet + HNet)\\
ENet &  Extraction network\\
D &  Discriminator\\
RNet &  Removal network\\
$a_i \in A$ &  To-be-processed image\\
$b_i \in B$ &  Processed unmarked image \\
$b_i'\in B'$&  Processed marked image\\
$b_i''\in B''$&  Output of RNet\\
$\xi$ &  Embedded watermark\\
$\xi_0$ &  Null watermark (all-white)\\
$\xi_\text{Overwrite}$ &  Watermark for overwriting\\
\hline
\hline
\end{tabular}}
\end{table}

\section{Problem Formulation}
\subsection{Image-to-Image Model Box-Free Watermarking}
\label{sec:image_to_image_model}

The flowchart of the existing box-free watermarking methods is presented in Fig. \ref{fig:victim_model} with image denoising as an example, and the list of main symbols and abbreviations is provided in Table \ref{tab:notation}. We assume that the original images $a_1, a_2, \ldots, a_M$ belong to domain $A$, and corresponding processed (denoised, in this example) images $b_{1}, b_2, \ldots b_{M}$ belong to domain $B$.
To protect the pretrained IPNet from model stealing attacks, HNet is employed to embed an invisible watermark, denoted by $\xi$ (the IEEE mark image in Fig. \ref{fig:victim_model} for example), into IPNet output $b_i$. Specifically, $b_{i}$ and $\xi$ are channel-wise concatenated and fed into HNet, which then yields the watermarked $b'_{i}$ in domain $B'$. Meanwhile, ENet is supposed to generate the watermark image from $b'_{i}$ as close to $\xi$ as possible and generate an all-white image, denoted by $\xi_0$, from non-watermarked images\footnote{The predefined ENet output images for non-watermarked inputs can be noise, solid colors, or other low entropic images. In this paper, we use an all-white image as shown in Fig. \ref{fig:victim_model}.}. To achieve these functionalities, with the pretrained IPNet frozen, HNet and ENet are jointly trained by minimizing
\begin{equation}\label{eq:victim_training_loss}
    \mathcal{L}_\text{Joint} = \alpha_1 {\mathcal{L}}_{\text{Hide}} + \alpha_2 {\mathcal{L}}_{\text{Mark}} + \alpha_3 {\mathcal{L}}_{\text{Clean}},
\end{equation}
where 
\begin{align}\label{eq:victim_losses}
    \mathcal{L}_\text{Hide} & = \frac{1}{N}\sum\nolimits_{b_i'\in {B}', b_i\in {B}} \| {b_i'} - {b_i} \|_2^2, \\
    \mathcal{L}_\text{Mark} &  = \frac{1}{N}\sum\nolimits_{b_i\in {B}'} \| \text{ENet}\left ( b_{i} \right )- \xi \|^2_2, \\
    \mathcal{L}_\text{Clean} & = \frac{1}{N}\sum\nolimits_{b_i\in {B}} \| \text{ENet}\left ( b_{i} \right )- \xi_{0} \|^2_2,
\end{align}
are the hiding loss, watermark loss, and clean loss, respectively, $\alpha_{1}$, $\alpha_{2}$, and $\alpha_{3}$ are weighting hyperparameters, and $N$ is the number of pixels. There exist a few optional loss functions, e.g., the perceptual loss\cite{johnson2016perceptual} and consistency loss \cite{zhang2020model} for some performance improvements, while $D$ can also be omitted. It has been show in \cite{wu2020watermarking} and \cite{zhang2021deep} that ENet can extract the hidden watermark image not only from $b'$s generated by HNet, but also from surrogate model outputs. This indicates that one can not only verify images generated by IPNet but also trace back surrogate models the approximate IPNet.

\begin{figure*}[!t]
  \centering
  \includegraphics[width=2\columnwidth]{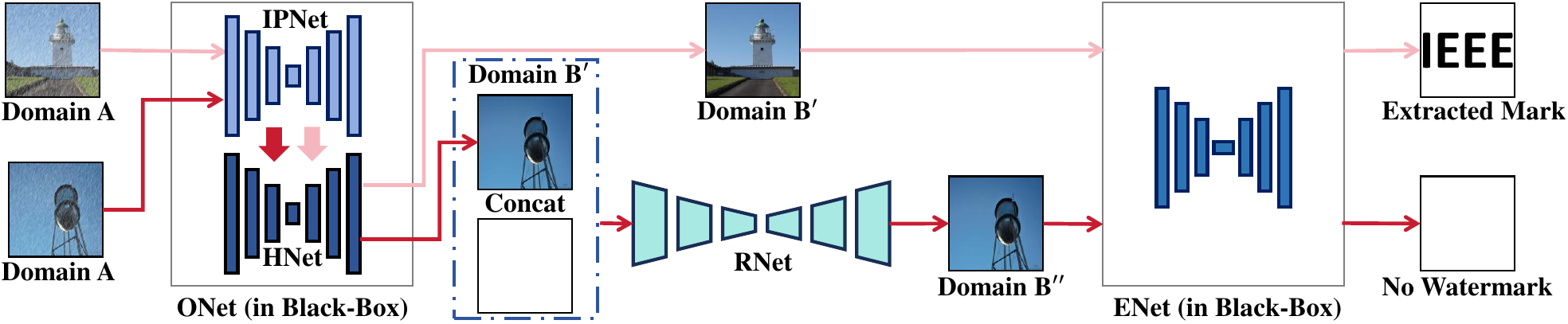}
  \vspace*{-6pt}
  \caption{Flowchart of the victim model and the proposed attack, where ONet and ENet are secured in black-box, with only their API available to attacker. Images processed by RNet will be watermark-free, corresponding to all-white outputs from ENet.}
  \label{fig:overall_attack}
  \vspace{-3mm}
\end{figure*}

\begin{table}[!h]
\renewcommand{\arraystretch}{1.0}
\centering
\caption{The three scenarios (Scen.) considered in this paper, in which all models are in black box. ``Activ.'' stands for activation.}
\label{tab:scenarios_diff} 
\vspace*{-9pt}
\setlength{\tabcolsep}{4pt}
{\small{\begin{tabular}{c|c|c|c|l}
\hline
\hline
Scen. & \begin{tabular}{@{}c@{}} IPNet \\ API \\ $A \to B$  \end{tabular} & 
        \begin{tabular}{@{}c@{}} HNet \\  API \\ $B \to B'$ \end{tabular} &
        \begin{tabular}{@{}c@{}} ONet \\  API \\ $A \to B'$ \end{tabular} &
        \multicolumn{1}{c}{ ENet API} \\
\hline
1 & \multirow{3}{*}{\ding{55}} &
    \multirow{3}{*}{\ding{55}} & 
    \multirow{3}{*}{\ding{51}} &  
    \ding{51}, ReLU Activ. Only \\
2 &  &  & & \ding{51}, Unknown Activ. \\
3 &  &  & & \ding{55}, Inaccessible \\
\hline
\hline
\end{tabular}}}
\end{table}

\subsection{Threat Model}
\label{sec:threat_model}
For the ease of discussion, we consider the victim model consisting of IPNet and HNet as a single entity called operation network (ONet). The three scenarios are then summarized in Table \ref{tab:scenarios_diff} that defines our threat model. Combining Fig. \ref{fig:victim_model} and Table \ref{tab:scenarios_diff}, we note that in all scenarios, the attacker can only observe the to-be-processed images in domain $A$ and the processed and marked images in domain $B'$, while the intermediate results ($b_i \in B$) are strictly private. The ReLU activation condition in Scenario 1 may not hold in practice, but this scenario is considered for theoretical analysis and is followed up by the more practical Scenario 2. In practice, Scenario 2 can be the case that ENet is provided as a black-box cloud service and can be used by end users to verify if an image is generated by IPNet, while Scenario 3 corresponds to the situation in which ENet is kept private by an authority for watermark verification.

For Scenarios 1 and 2, the attacker designs the RNet such that 1) if RNet input is a watermarked image, then the output is the watermark-removed image, and 2) if RNet input is a non-watermarked image, then the output does not contain the watermark, either. In both cases, the output image quality should be close to that of the input. For Scenario 3, since the observable information is minimal, we resort to transferability and design a proxy remover denoted by RNet$'$, which is then used in the same way as RNet.

\begin{figure*}[t]
  \centering
  \includegraphics[width=2\columnwidth]{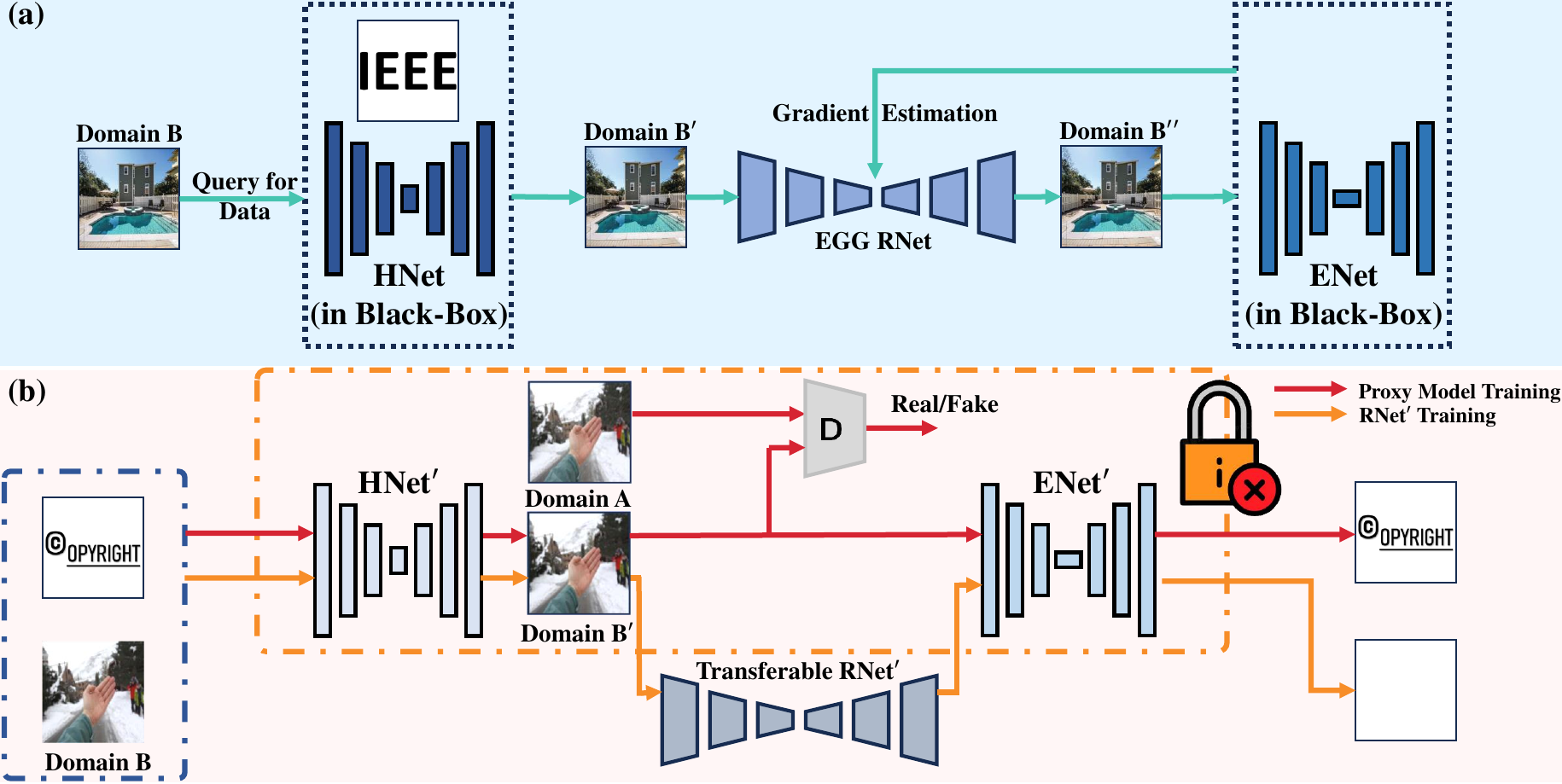}
    \caption{Flowchart of the training process for the proposed (a) EGG RNet and (b) Transferable RNet. Note that the proxy models are private and are frozen during the training of the Transferable RNet.}
  \label{fig:overall_training}
  \vspace{-3mm}
\end{figure*}

\section{Proposed Attacks}
The general flowchart of the proposed attack against the victim model is shown in Fig. \ref{fig:overall_attack}. while the flowcharts of the training process for EGG RNet and Transferable RNet are presented in Fig. \ref{fig:overall_training}. IPNet in Fig. \ref{fig:overall_training} is omitted for brevity. We note that querying HNet is required in the EGG attack but not in the transferable attack. The components that are used to perform the attack are the EGG RNet and Transferable RNet, respectively. For all scenarios, RNet takes an image $b_{i}'$ from domain $B'$, concatenated with the all-white image $\xi_0$ as input, and attempts to generate the watermark-removed output $b_{i}''$. For a non-watermarked input, RNet will yield a high quality non-watermarked output. This is modeled by
\begin{equation}
\label{eq:concatenate}
b_i''=\text{RNet}(b_i' \oplus \xi_0),
\end{equation}
where $\oplus$ denotes channel-wise concatenation. In what follows, we first present the EGG removers under Scenarios 1 and 2, and then the transferable remover under Scenario 3.

\subsection{Scenarios 1 and 2: EGG Removers}
\label{sec:observable-extractor-guided remover}
Since ONet performs two tasks via IPNet and HNet on a query image, its output $b_i'$ is then the processed and watermarked image. An ideal removal attack should compensate HNet only, while keeping the image alteration due to IPNet intact. Unfortunately, the intermediate result $b_i$ is inaccessible, making it impossible to reverse the functionality of HNet by observing its input ($b_i$)-output ($b_i'$) pairs. Under this stringent but practical attack condition, we are instead interested in exploring whether a self-supervised image reconstructor, supervised only by the observable $b_i'$, can do the job. Therefore, in Scenarios 1 and 2, we minimize the following loss function when training RNet,
\begin{equation}\label{eq:loss_objective}
    \mathcal{L}_\text{RNet} = \frac{1}{N} \sum_{b_{i}''\in B'', b_{i}'\in B'} \left ( \beta_{1} \mathcal{L} _{\text{Quality}} + \beta_{2}\mathcal{L} _{\text{Remove}} \right ),
\end{equation}
where
\begin{align}
\mathcal{L}_{\text{Quality}} & = \left \| b''_{i}-b'_{i} \right \|^{2}_{2}, \label{eq:loss_quality}\\
\mathcal{L}_{\text{Remove}} &  = \left \| \text{ENet}\left ( b''_{i}  \right ) -\xi_{0} \right \| ^{2}_{2}, \label{eq:loss_wm}
\end{align}
$\beta_{1}$ and $\beta_{2}$ are weighting hyperparameters. $\mathcal{L} _{\text{Quality}}$ preserves the image quality after watermark removal, while $\mathcal{L} _{\text{Remove}}$ ensures that the output $b_i''$ does not contain the watermark. Note that (\ref{eq:loss_quality}) is practically obtainable since it only depends on the observable $b_i'$ from ONet API. Unfortunately, the problem arises when evaluating (\ref{eq:loss_wm}) which involves querying ENet. Specifically, it requires the gradients of ENet to be back-propagated during training of RNet. However, ENet is in a black-box, and only its output is available. 

\begin{figure}[!h]
  \centering
  \includegraphics[width=0.4\textwidth]{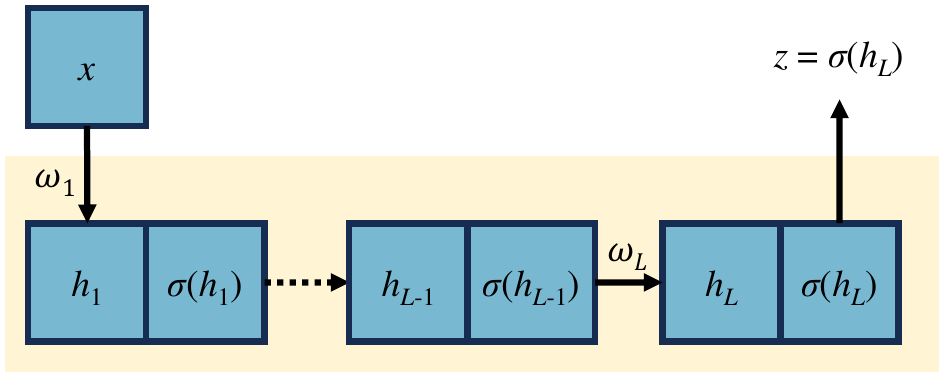}
  \vspace*{-6pt}
  \caption{Block diagram of a simplified ENet.}
  \label{fig:simple_network}
  \vspace*{-6pt}
\end{figure}

\subsubsection{Scenario 1: ENet Has ReLU Activation Only}
Without loss of generality and for mathematical tractability, we consider a simplified ENet depicted in Fig. \ref{fig:simple_network}, where $x$ is the input\footnote{Note that the query $x$ to ENet can be a watermarked image, i.e., $b_i'$, or a watermark-free image from any domains, e.g., $A$, $B$, open set, or watermark-removed domain $B''$.}, $\omega_i$ the weight, $h_i$ the logits, $\sigma$ the activation function, and $z$ the output ($\text{ENet}(x)$). To update ENet's parameters, ${\partial \text{Loss}}/{\partial x}$ must be independent of the elements within ENet, including $\omega_{i}$, $h_i$, and $\sigma$, $1 \leq i \leq L$. Applying the chain rule, we have
\begin{equation}\label{eq:backpropagation}
\frac{\partial \text{Loss}}{\partial x} = \frac{\partial \text{Loss}}{\partial z}\prod_{i=1} ^{L} \frac{\partial \sigma\left ( h_{i} \right ) }{\partial h_{i} }w_{i},
\end{equation}
where Loss is the adversary-defined loss function, e.g., (\ref{eq:loss_wm}). If ENet only uses ReLU for activation, then (\ref{eq:backpropagation}) simplifies to
\begin{equation}
\label{eq:backpropagation_with_relu}
{\left. {\frac{{\partial {\text{Loss}}}}{{\partial x}}} \right|_{{\text{ReLU}}}} = \begin{cases}
 \frac{\partial \text{Loss}}{\partial z} \prod_{i=1}^{L}w_{i} ,&  \forall h_{i} > 0, \\ 
{0,}& {\text{otherwise,}} 
\end{cases}
\end{equation}
and the forward propagation yields 
\begin{equation}
\label{eq:forward_with_relu_z}
z = \begin{cases}
\prod_{i=1}^{L} w_i x ,& x \text{ and } \forall h_i > 0, \\ 
{0,} & {\text{otherwise.}} 
\end{cases}
\end{equation}
Substituting (\ref{eq:forward_with_relu_z}) into (\ref{eq:backpropagation_with_relu}), we have 
\begin{equation}
\label{eq:backpropagation_with_relu_simplified}
{\left. {\frac{{\partial {\text{Loss}}}}{{\partial x}}} \right|_{{\text{ReLU}}}} = \begin{cases}
 \frac{\partial \text{Loss}}{\partial z}\frac{z}{x},& x \text{ and } \forall h_i > 0, \\ 
{0,}&{\text{otherwise.}} 
\end{cases}
\end{equation}
It can be seen from (\ref{eq:backpropagation_with_relu_simplified}) that ${\partial \text{Loss}}/{\partial x}$ is a function of the adversary-defined loss, input $x$, and output $z$, which are all available. This shows the feasibility of the EGG RNet based on querying the black-box ENet. 

\subsubsection{Scenario 2: ENet is Unknown}
In a more general and practical situation, the activation functions in ENet are unknown and can be of different forms, making the derivations in Scenario 1 invalid. Fortunately, methods to estimate the gradient information back-propagated from a black-box model output to its input have been extensively studied on the subject of black-box adversarial attacks \cite{chen2017zoo, dong2021query, shi2022query}. Accordingly, the estimated gradient is given by
\begin{equation}\label{eq:gradient_estimation}
    {\left. {\frac{{\partial {\text{Loss}}}}{{\partial x}}} \right|_{{\text{General}}}} \approx \sum\limits_i\frac{\text{ENet}\left ( {b}''+\delta {v}_i \right ) -\text{ENet}\left ( {b}'' \right ) }{\delta}{v}_i.
\end{equation}
where ${v}_i$ are a set of direction vectors typically Gaussian and with unit norm. Great research efforts have been devoted to more efficient estimations in (\ref{eq:gradient_estimation}), e.g., reducing the number of queries while preserving the estimation accuracy, e.g., \cite{dong2021query,shi2022query}. With the above results, the training of the EGG RNet supervised by (\ref{eq:loss_objective})--(\ref{eq:loss_wm}) can be effectively carried out.

\subsection{Scenario 3: Transferable Remover}
For Scenario 3, the attack strategy exploits transferability. Since the functionality of IPNet is known, we can design and train a set of proxy models denoted by IPNet$'$, HNet$'$, and ENet$'$, respectively. Their architectures can be different from those of IPNet, HNet, and ENet, but the targeted functionalities are the same. Then, we freeze these proxy models and train RNet$'$ in a white-box sense  minimizing the following loss function
\begin{equation}\label{eq:transfer_loss_objective}
    \mathcal{L}_{\text{RNet}'}= \frac{1}{N} \sum_{s_{i}''\in S'', s_{i}'\in S'}\left ( \beta_{3} \mathcal{{L}} _{\text{Quality}'} + \beta_{4} \mathcal{{L}} _{\text{Remove}'} \right ),
\end{equation}
where, similar to (\ref{eq:loss_quality}) and (\ref{eq:loss_wm}),
\begin{align}
    \mathcal{{L}} _{\text{Quality}'} & = \left \| s''_{i}-s'_{i} \right \|^{2}_{2}, \label{eq:transfer_loss_quality}\\ 
    \mathcal{{L}} _{\text{Remove}'} & = \left \| \text{ENet}'\left ( s''_{i}  \right ) -\xi_{0} \right \| ^{2}_{2},\label{eq:transfer_loss_wm}
\end{align}
$\beta_{3}$ and $\beta_{4}$ are weights, $s_i'$ and $s_i''$ correspond to $b_i'$ and $b_i''$, respectively, but are from our private dataset, $S'\subset B'$, and $S''\subset B''$, respectively. We make the following attempt to explain why transferability holds in the attack.

Watermark embedding and extraction can be considered as a coupling and decoupling function, respectively, between the carrier image and the watermark, either in the conventional rule-based or the recent deep-learning-based approach. This process, in the context of box-free model watermarking, is the following flow,
\begin{equation}\label{flow:victim_model_workflow}
    b_i \to \text{HNet}\left (b_i, \xi  \right ) = b_{i}' \to \text{ENet}\left (  b_{i}'\right ) = \xi,
\end{equation}
which indicates that HNet is the coupler while ENet is the decoupler. The underlying mappings learned by these two models can thus also be approximated by our proposed proxy models via appropriate supervision, and this is achieved by establishing the following chain using our private dataset and proxy models, i.e.,  
\begin{equation}\label{eq:mapping_workflow}
    s_i \to {\text{HNet}'}\left (s_i, \xi  \right ) = s_{i}' \to \text{ENet}'\left (  s_{i}'\right ) = \xi,
\end{equation}
regardless of the chosen model architectures, and as long as $S'$ and $S''$ have overlap with $B'$ and $B''$, respectively, which is available in our practical threat model. Therefore, the remover created against chain (\ref{eq:mapping_workflow}) can be transferred to attack chain (\ref{flow:victim_model_workflow}), which supports our hypothesis.

\section{Experimental Results}
\label{sec:Experimental_Results}
To demonstrate the effectiveness of our proposed methods, we launch the proposed removers against the state-of-the-art box-free model watermarking methods, i.e., \cite{wu2020watermarking, zhang2021deep}. We note that box-free model watermarking is applicable across various models including natural language processing, reinforcement learning, and generative models such as diffusion models, etc. In this paper, following the existing works, we focus on the image processing models mentioned in the victims and image deraining is considered as the underlying task. For convenience, we denote the victim model in \cite{wu2020watermarking} by ``$\text{V}1$'' and the one in \cite{zhang2021deep} by ``$\text{V}2$''. The experimental results are based on our own implementation of V1 \cite{wu2020watermarking} and the official implementation of V2 \cite{zhang2021deep}. For the generalizability test, we will launch the proposed transferable removal attack against \cite{huang2023can} which only shares the same verification process, and we denote the victim model in \cite{huang2023can} by ``$\text{V}3$''.  We first provide the details of our experimental settings, followed by qualitative and quantitative experimental results.

\subsection{Experimental Settings}
\label{sec:experimental_details}
\subsubsection{Dataset}
\label{sec:dataset_setup}
We use the PASCAL VOC dataset \cite{everingham2010pascal} and consider the image deraining task. Its training set comprises $12,000$ noisy images from domain $A$ and $12,000$ denoised images from domain $B$. To reflect the real-world situation, we divide the data into two disjoint equal halves, each consisting of $6,000$ images, so that the training data for the victim model and the removers are mutually exclusive. To reduce the computational complexity, we downscale the original $512 \times 512$ RGB images to $256 \times 256$ grayscale images.

\begin{table}[!h]
  \centering
  \caption{Summary of implemented model architectures}
  \renewcommand{\arraystretch}{1.0}
  \label{tab:architectures}
  \vspace*{-9pt}
  \resizebox{0.9\columnwidth}{!}{
  \begin{tabular}{c|c|ccc}
  \hline
  \hline
  & \textbf{Model} & \textbf{HNet} & \textbf{ENet} & \textbf{RNet} \\ \hline
  \multirow{2}{*}{\textbf{Victim}} & V1 \cite{wu2020watermarking} & UNet & EEENet &  -- \\
  & V2 \cite{zhang2021deep} & UNet & CEILNet & --\\ 
  \hline
  \multirow{5}{*}{\textbf{Attack}} & EGG & -- & -- & UNet \\
  \cline{2-5}
  & \multirow{4}{*}{Transferable} & \textbf{HNet}$'$ & \textbf{ENet}$'$ & \textbf{RNet}$'$ \\
  \cline{3-5}
  & & \begin{tabular}{@{}c@{}} CGAN \\ ResNet50 \\ UNet  \end{tabular} 
               & \begin{tabular}{@{}c@{}} EEENet \\ CEILNet\end{tabular} 
               & UNet \\
  \hline
  \hline 
  \end{tabular}}
  \vspace*{-9pt}
  \end{table}

\subsubsection{Model}
\label{sec:model_details}
The victim models in \cite{wu2020watermarking} and \cite{zhang2021deep} share the same framework depicted in Fig. \ref{fig:victim_model}, and both use the UNet \cite{ronneberger2015u} as the hiding network HNet. In our experiments, we further assume that both victim models use the same IPNet since the image processing functionality is unrelated to the watermarking scheme. However, the two victim models use different ENet architectures. In \cite{wu2020watermarking}, the ENet was newly designed as a concatenation of three subnets, and we called it EEENet ($0.1$ M). EEENet comprises three sub-networks: $E_A$, $E_B$, and $E_C$. Specifically, $E_A$ receives the watermarked image as input, and $E_B$ receives the key as input. The outputs of both $E_A$ and $E_B$ are added together and used as the input to $E_C$, where the extracted watermark is obtained at the output of $E_C$. In \cite{zhang2021deep}, the authors incorporated the CEILNet \cite{fan2017generic} ($3.4$ M) as the ENet. In our attacks, the UNet architecture is implemented for RNet. In Scenario 3, since the victim hiding network is unknown, we implement the proxy hiding network $\text{HNet}'$ by selecting one model from CGAN\cite{isola2017image} ($5.5$ M), ResNet50 \cite{he2016deep} ($22.8$ M), and UNet ($41.8$ M) \cite{ronneberger2015u}, all commonly used in the computer vision field. Additionally, the proxy extraction network $\text{ENet}'$ is selected between CEILNet and EEENet. The implemented models are summarized in Table \ref{tab:architectures}.

\subsubsection{Training Detail}
\label{sec:training_details}
All models are trained from scratch for $100$ epochs using the Adam optimizer, with an initial learning rate of $0.0002$. For the training of different architectures for $\text{HNet}'$ and $\text{ENet}'$, we adopt the loss functions and hyperparameter settings in \cite{zhang2021deep}. Specifically, for (\ref{eq:victim_training_loss}), we set $\alpha_{1}=\alpha_{2}=\alpha_{3}=10^4$, and such a boost is required to ensure the balance with $\mathcal{L}_\text{Task}$. A similar setup is followed for the training of the proposed removers. For (\ref{eq:loss_objective}) and (\ref{eq:transfer_loss_objective}), we set $\beta_{1}=\beta_{2}=\beta_{3}=\beta_{4}=10^4$. Since Scenario 1 can be considered as a special case of Scenario 2 where ENet only uses ReLU activation, for brevity, we only provide experimental results in the more practical Scenarios 2 (EGG RNet) and 3 (transferable RNet$'$).
\begin{figure*}[!t]
  \centering
  \includegraphics[width=2.0\columnwidth]{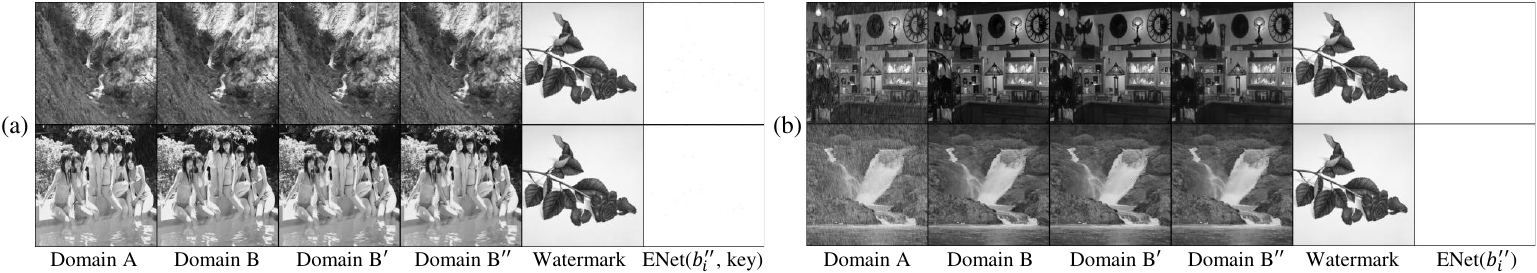}
  \vspace*{-9pt}
  \caption{Qualitative demonstration of the proposed EGG watermark remover RNet against victim models (a) V1 \cite{wu2020watermarking} and (b) V2 \cite{zhang2021deep}.}
  \vspace*{-6pt}
  \label{fig:qualitative_result4}
\end{figure*}

\begin{figure*}[!t]
  \centering
  \includegraphics[width=1.9\columnwidth]{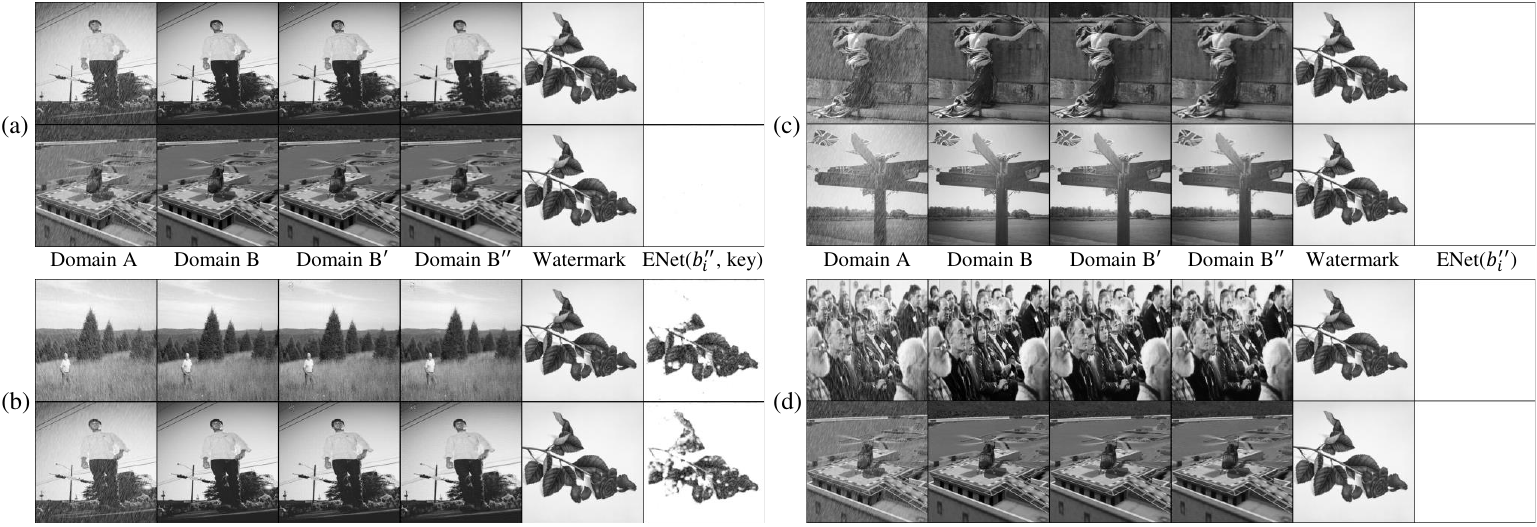}
  \vspace*{-9pt}
  \caption{Qualitative demonstration of the proposed transferable remover RNet$'$ against the victim models, where HNet$'$ is CGAN. (a) ENet$'$ is CEILNet, against V1 \cite{wu2020watermarking}. (b) ENet$'$ is EEENet, against V1 \cite{wu2020watermarking}. (c) ENet$'$ is CEILNet, against V2 \cite{zhang2021deep}. (d) ENet$'$ is EEENet, against V2 \cite{zhang2021deep}.}
  \label{fig:qualitative_result1}.
  \vspace*{-3pt}
\end{figure*}

\subsubsection{Performance Metric}
\label{sec:performance_metrics}

We use the peak signal-to-noise ratio (PSNR) and multi-scale structural similarity index (MS-SSIM) \cite{wang2003multiscale} to evaluate the image quality. Given a set of watermarked images, we defined 1) the success rate of watermark extraction, i.e., SR$_\text{E}$, as the ratio of the images whose watermarks are successfully extracted by ENet \cite{zhang2021deep}, 2) the success rate of watermark removal, i.e., SR$_\text{R}$, as the ratio of the images, after being processed by RNet, whose corresponding ENet outputs become an all-white image, and additionally 3) the success rate of watermark overwrite, i.e., SR$_\text{O}$, as the success rate of overwritten watermark extraction from ENet when the watermarks are overwritten by the modified EGG remover. For the three success rates, a successful extraction is confirmed if the normalized correlation coefficient between the intended ENet output (watermark or an all-white image) and the actual ENet output is greater than $0.96$. 

\begin{figure*}[!t]
  \centering
  \includegraphics[width=1.9\columnwidth]{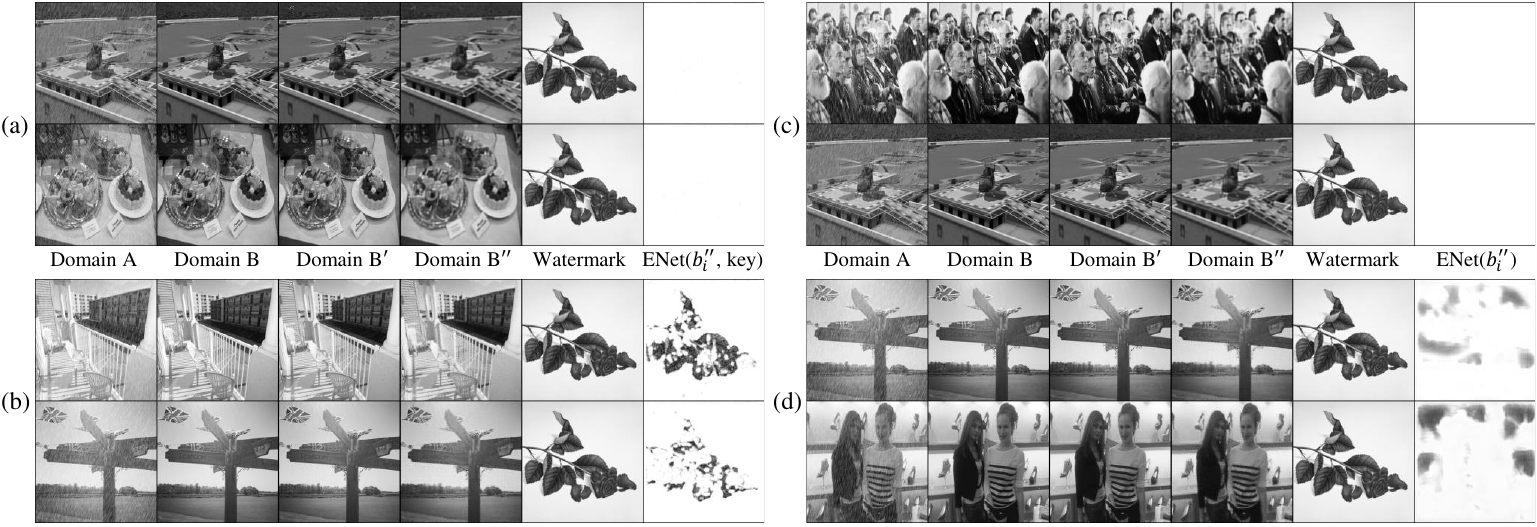}
  \vspace*{-9pt}
  \caption{Qualitative demonstration of the proposed transferable remover RNet$'$ against the victim models, where HNet$'$ is ResNet50. (a) ENet$'$ is CEILNet, against V1 \cite{wu2020watermarking}. (b) ENet$'$ is EEENet, against V1 \cite{wu2020watermarking}. (c) ENet$'$ is CEILNet, against V2 \cite{zhang2021deep}. (d) ENet$'$ is EEENet, against V2 \cite{zhang2021deep}.}
  \label{fig:qualitative_result2}
  \vspace*{-3pt}
\end{figure*}

\begin{figure*}[!t]
  \centering
  \includegraphics[width=1.9\columnwidth]{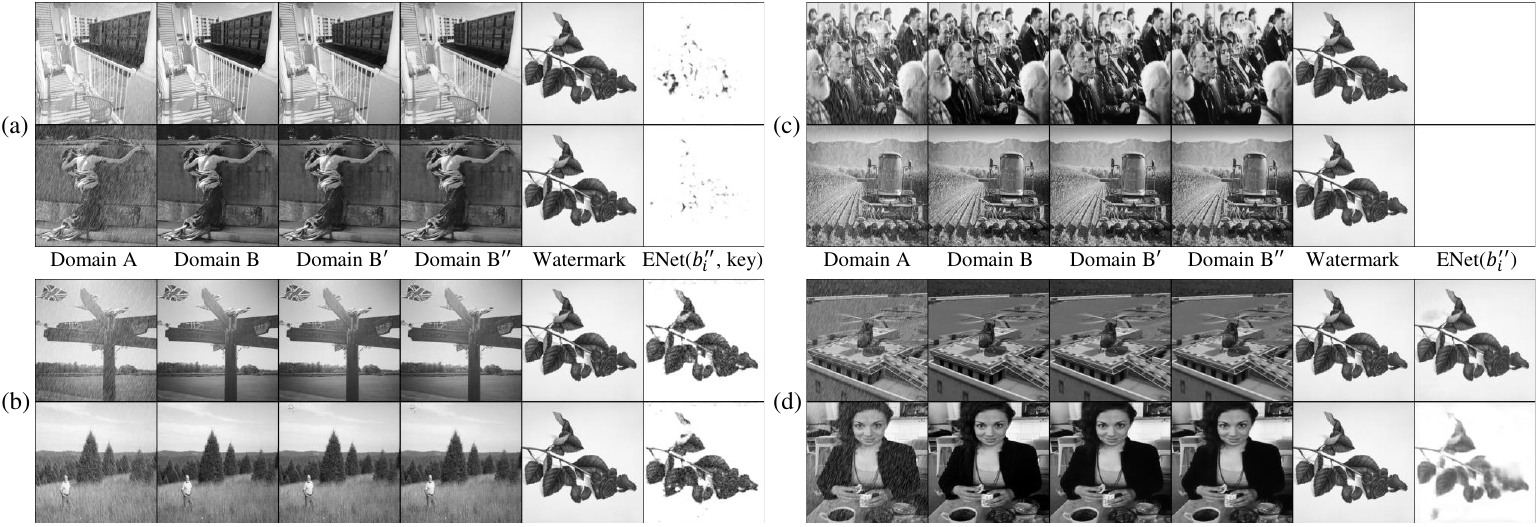}
  \vspace*{-9pt}
  \caption{Qualitative demonstration of the proposed transferable remover RNet$'$ against the victim models, where HNet$'$ is UNet. (a) ENet$'$ is CEILNet, against V1 \cite{wu2020watermarking}. (b) ENet$'$ is EEENet, against V1 \cite{wu2020watermarking}. (c) ENet$'$ is CEILNet, against V2 \cite{zhang2021deep}. (d) ENet$'$ is EEENet, against V2 \cite{zhang2021deep}.}
  \label{fig:qualitative_result3}
  \vspace*{-3pt}
\end{figure*}

\subsection{Qualitative Evaluation}
\label{sec:qualitative evaluation}

\subsubsection{EGG Remover}
The demonstrative examples of the proposed EGG removal attack against the two victims models V1 \cite{wu2020watermarking} and V2 \cite{zhang2021deep} are provided in Fig. \ref{fig:qualitative_result4} (a) and (b), respectively, where the columns represent the to-be-processed image (domain $A$), processed image (domain $B$), processed and watermarked image (domain $B'$), processed and watermark-removed image (domain $B''$), original watermark image, and victim ENet extracted output, respectively. Note that in V1 \cite{wu2020watermarking}, the extraction requires the correct key as an input to its ENet. The extraction result is thus denoted by $\text{ENet}(b_i'', key)$, and we used the correct key in the experiments to verify the effectiveness of the remover. It can be seen from the figure that the proposed EGG RNet can successfully remove the watermark and turn it into an almost all-white image.

\subsubsection{Transferable Remover}
\label{sec:Attack Results for Transferable Remover}
To achieve optimal attack performance against a specific victim, the proposed transferable attack involves training a set of proxy models for testing. This is necessary due to the attacker's lack of prior knowledge about the victim in the black-box setting. Based on the choice of the architecture of proxy hiding net HNet$'$, the demonstrative examples are shown in Figs. \ref{fig:qualitative_result1}, \ref{fig:qualitative_result2}, and \ref{fig:qualitative_result3}, where HNet' is CGAN, ResNet50, and UNet, respectively, and ENet' is selected between CEILNet and EEENet. Among the subfigures, (a) and (b) show attacks against V1 \cite{wu2020watermarking}, while (c) and (d) show attacks against V2 \cite{zhang2021deep}. We observe that V2 is more vulnerable than V1, e.g., in Fig. \ref{fig:qualitative_result1}, after the attack, the extracted watermark is completely invisible in (d), while in (b), the extracted watermark's outline remains visible, albeit less clear. A similar phenomenon can also be observed between Fig. \ref{fig:qualitative_result3} (a) and (c). We hypothesize that the reason lies in the additional key authentication step in the watermark extraction in V1. Nevertheless, the transferable remover remains effective in attacking V1, e.g., the extraction results in Figs. \ref{fig:qualitative_result1} (a) and \ref{fig:qualitative_result2} (a) successfully had the watermarks removed. We further observe that the effectiveness of the remover trained with different $\text{HNet}'$ architectures varies against the two victim models. According to the subfigure pairs of (b) and (d) across Figs. \ref{fig:qualitative_result1}, \ref{fig:qualitative_result2}, and \ref{fig:qualitative_result3}, it is evident that CGAN demonstrates superior performance over ResNet50 followed by UNet. 
By comparing subfigures (a) and (b), or (c) and (d), it can be seen that CEILNet outperforms EEENet for the implementation of $\text{ENet}'$.

\begin{figure}[!t]
  \centering
  \includegraphics[width=0.95\columnwidth]{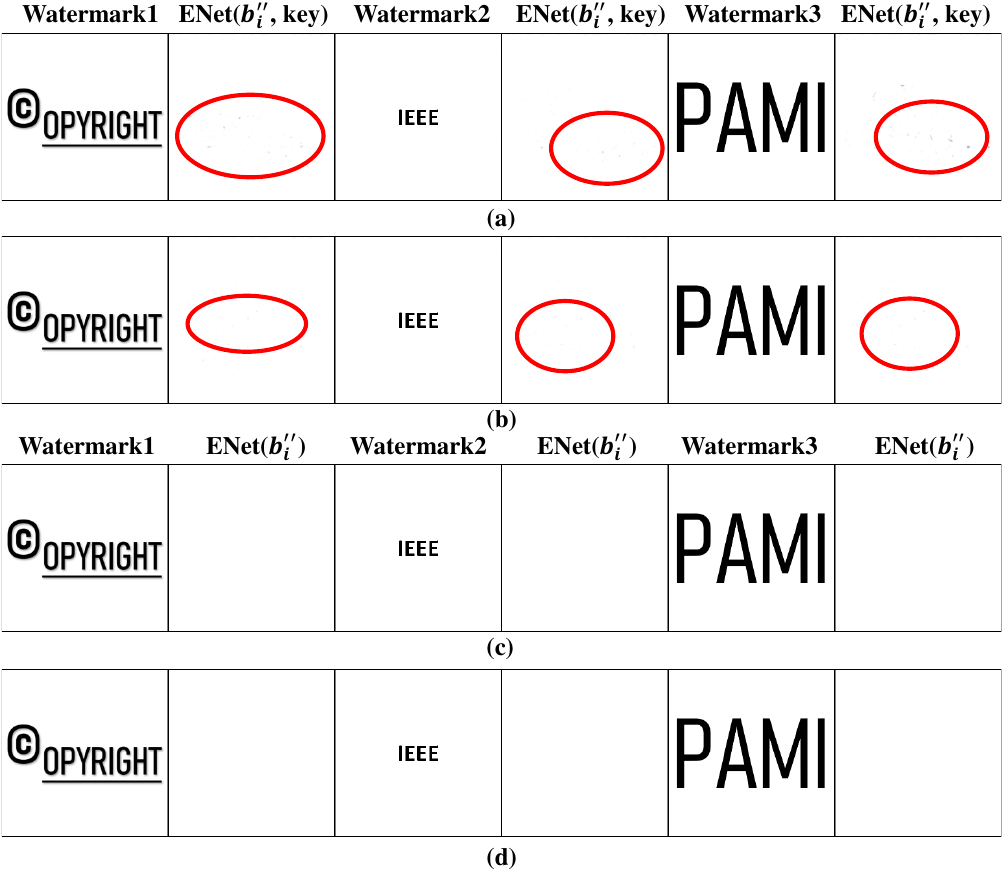}
  \vspace*{-9pt}
  \caption{Demonstration of watermark sensitivity of the proposed attacks. (a) EGG remover against V1 \cite{wu2020watermarking}. (b) Transferable remover against  V1 \cite{wu2020watermarking}. (c) EGG remover against V2 \cite{zhang2021deep}. (d) Transferable remover against V2 \cite{zhang2021deep}. The red circles highlight the non-white spots in the extraction results, which do not affect the correlation coefficients.}
  \label{fig:different_secret_attack}
  \vspace*{-9pt}
\end{figure}

\begin{figure}[!t]
  \centering
  \includegraphics[width=1\columnwidth]{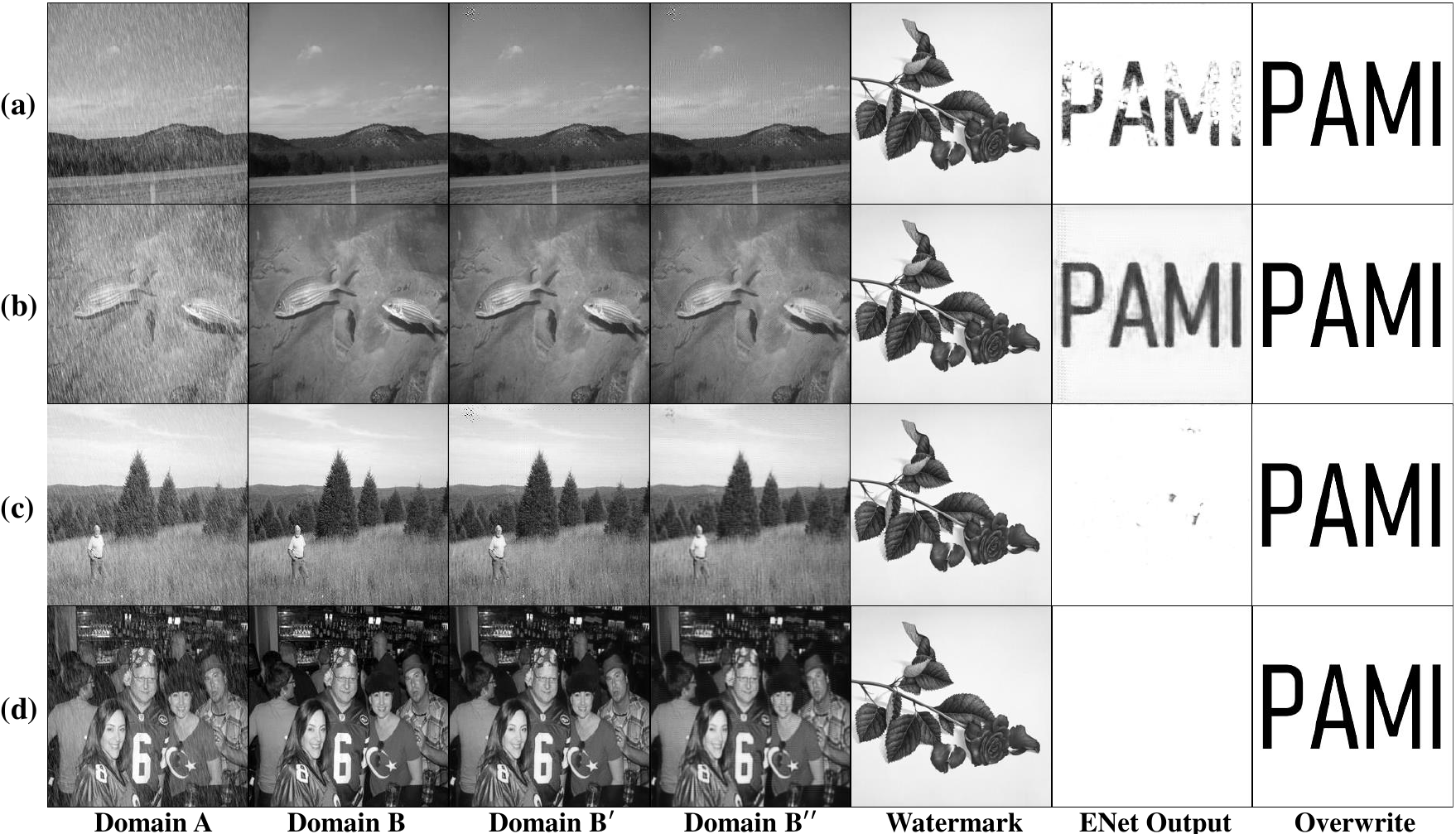}
  \vspace*{-18pt}
  \caption{Demonstrative examples of watermark overwriting. (a) EGG remover against V1 \cite{wu2020watermarking}, with correlation coefficient of 0.961. (b) EGG remover against V2 \cite{zhang2021deep}, with correlation coefficient of 0.984. (c) Transferable remover (CGAN plus CEILNet) against V1 \cite{wu2020watermarking}. (d) Transferable remover (CGAN plus CEILNet) against V2 \cite{zhang2021deep}.}
  \label{fig:replace_wm}
  \vspace{-6pt}
\end{figure}

\subsubsection{Watermark Sensitivity}
\label{sec:Attacks on Different Embedded Watermarks}
We now examine the performance of the proposed attacks in terms of the sensitivity to different watermark images, and the results are shown in Fig. \ref{fig:different_secret_attack}, where the transferable remover is realized by CGAN and CEILNet. By comparing (a) with (c), and (b) with (d), we observe a consistent finding that V2 \cite{zhang2021deep} is more vulnerable than V1 \cite{wu2020watermarking}. There exist non-white spots in the extracted images from V1 \cite{wu2020watermarking}, but they have minimal impact on the correlation coefficients. Overall, Fig. \ref{fig:different_secret_attack} confirms that the proposed attacks are effective against different types of watermark images.

\begin{table}[!t]
  \centering
  \caption{Victim model watermark robustness test against JPEG compression, where PSNR is in dB, $0< \text{MS-SSIM} <1$, and $0<\text{SR}_{\text{E}}<1$.}
  \vspace*{-9pt}
  \setlength{\tabcolsep}{2pt}
  \label{tab:JPEG_compression}
  \begin{tabular}{c|ccc|ccc}
  \hline
  \hline
  \multirow{2}{*}{Factor} & \multicolumn{3}{c}{V1 \cite{wu2020watermarking}} & \multicolumn{3}{c}{V2 \cite{zhang2021deep}} \\
  \cline{2-7}
  & PSNR$\uparrow$ & MS-SSIM$\uparrow$ & $\text{SR}_{\text{E}}\uparrow$ & PSNR$\uparrow$ & MS-SSIM$\uparrow$ & $\text{SR}_{\text{E}}\uparrow$  \\
  \hline
  $50\%$ & $30.22$ & $0.9873$ & $0.10$ & $30.99$ & $0.9875$ & $0.00$ \\ 
  $60\%$ & $31.30$ & $0.9909$ & $0.20$ & $32.01$ & $0.9905$ & $0.00$ \\ 
  $70\%$ & $32.46$ & $0.9933$ & $0.72$ & $33.66$ & $0.9935$ & $0.00$ \\ 
  $80\%$ & $32.53$ & $0.9931$ & $1.00$ & $34.46$ & $0.9944$ & $0.00$ \\ 
  $90\%$ & $32.33$ & $0.9930$ & $1.00$ & $34.20$ & $0.9944$ & $0.00$ \\ 
  \hline
  \hline
  \end{tabular}
  \end{table}
  
\begin{table}[!t]
  \centering
  \caption{Victim model watermark robustness test against noise addition, where noise level and PSNR are in dB, $0< \text{MS-SSIM} <1$, and $0<\text{SR}_{\text{E}}<1$.}
  \vspace*{-9pt}
  \setlength{\tabcolsep}{2pt}
  \label{tab:Gaussian_noise}
  \begin{tabular}{c|ccc|ccc}
  \hline
  \hline
  \multirow{2}{*}{\begin{tabular}{@{}c@{}} Noise \\ Level \end{tabular}} & \multicolumn{3}{c}{V1 \cite{wu2020watermarking}} & \multicolumn{3}{c}{V2 \cite{zhang2021deep}} \\
  \cline{2-7}
  & PSNR$\uparrow$ & MS-SSIM$\uparrow$ & $\text{SR}_{\text{E}}\uparrow$ & PSNR$\uparrow$ & MS-SSIM$\uparrow$ & $\text{SR}_{\text{E}}\uparrow$  \\
  \hline
  $0$  & $9.38$ & $0.4557$ & $0.00$ & $9.36$ & $0.4551$ & $0.00$ \\
  $10$ & $16.80$ & $0.7539$ & $0.00$ & $16.79$ & $0.7523$ & $0.00$ \\
  $20$ & $25.39$ & $0.9298$ & $0.88$ & $25.57$ & $0.9288$ & $0.00$ \\ 
  $30$ & $30.90$ & $0.9838$ & $1.00$ & $31.80$ & $0.9846$ & $0.40$ \\
  $40$ & $32.34$ & $0.9920$ & $1.00$ & $33.86$ & $0.9935$ & $1.00$ \\
  \hline
  \hline
  \end{tabular}
\end{table}

\subsubsection{Watermark Overwriting}
\label{sec:overwrite}
The EGG remover is originally designed for watermark removal only, while in our experiments, we discovered that it can also be used to overwrite the watermark. For the completeness of our experiments, we make a slight modification to both the EGG and transferable removers to achieve watermark overwriting. Suppose that we attempt to use the new watermark $\xi_\text{Overwrite}$ to overwrite the unknown original watermark $\xi$, then (\ref{eq:concatenate}) modifies to
\begin{equation}
  \label{eq:overwrite}
  b_i''=\text{RNet}(b_i' \oplus \xi_\text{Overwrite}),
\end{equation}
while $\mathcal{L}_\text{Remove}$ in (\ref{eq:loss_wm}) and $\mathcal{L}_{\text{Remove}'}$ in (\ref{eq:transfer_loss_wm}), are replaced by
\begin{align}
  \mathcal{L}_\text{Overwrite} & = \left \| \text{ENet}\left ( b''_{i}  \right ) - \xi_\text{Overwrite} \right \| ^{2}_{2},\\
  \mathcal{L}_{\text{Overwrite}'} & = \left \| \text{ENet}'\left ( s''_{i}  \right ) - \xi_\text{Overwrite} \right \| ^{2}_{2},
\end{align}
respectively. With the above modification, the corresponding watermark overwriting experimental results are shown in Fig. \ref{fig:replace_wm}. It can be seen from Fig. \ref{fig:replace_wm} (a) and (b) that the proposed EGG remover can successfully overwrite the watermark against both victim models (correlation coefficients of $0.961$ and $0.984$, respectively), while the transferable remover failed to do so. Instead, the transferable overwriter destroys both the original and overwriting watermarks. We hypothesize that the reason lies in the conflicts among the victim hiding network, proxy hiding network, and overwriter.  

\begin{figure}[!t]
  \centering
  \includegraphics[width=1\columnwidth]{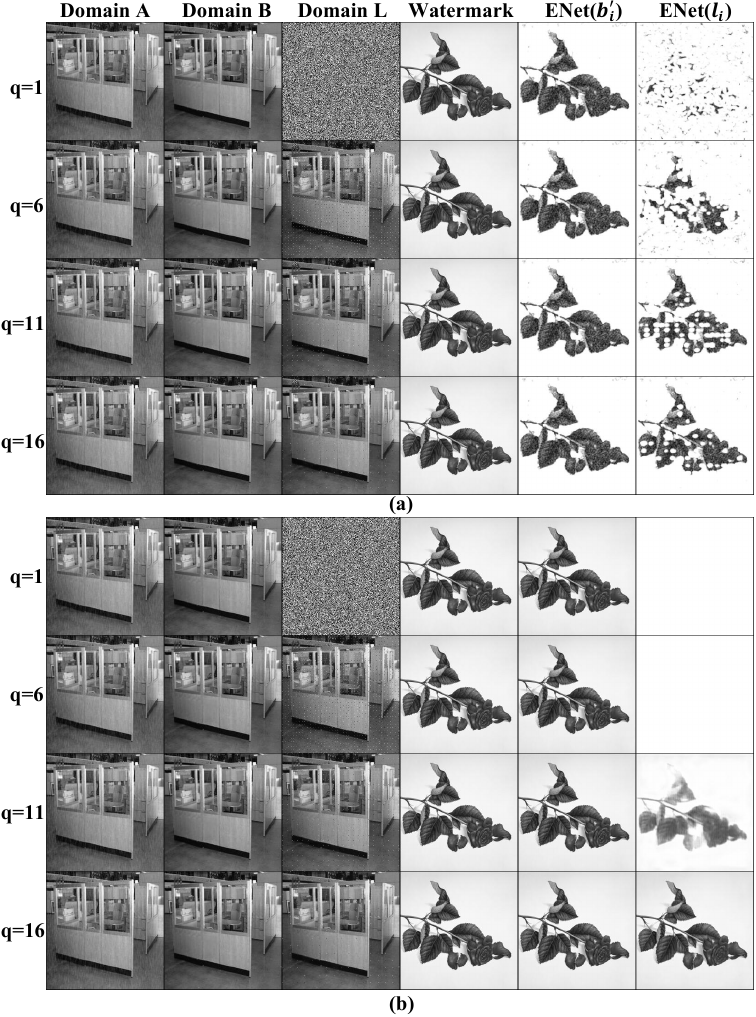}
  \vspace*{-18pt}
  \caption{Demonstrative examples of the lattice attack presented in \cite{liu2023erase} with various interval value q, where images $l_i$ in Domain $L$ are referred to as $b_i'$ which suffer from the lattice attack. (a) Lattice attack against V1 \cite{wu2020watermarking}. (b) Lattice attack against V2 \cite{zhang2021deep}.}
  \label{fig:lattice_attack}
  \vspace{-6pt}
\end{figure}

\begin{table*}[!t]
  \centering
  \caption{Quantitative evaluation results for proposed attacks, where PSNR is in dB, $0< \text{MS-SSIM} <1$, and $0<\text{SR}<1$. If the ``Target Extraction'' is ``Blank'', then the last column uses $\text{SR}_\text{R}$ to test watermark removal. Otherwise, the last column uses $\text{SR}_\text{O}$ to test watermark overwriting.}
  \vspace*{-6pt}
  \label{tab:quantitative_result} 
  \resizebox{1\textwidth}{!}{
  \begin{tabular}{c|c|c|c|c|c|c|c|c|c|c|c}
  \hline
  \hline
  Index & \begin{tabular}[c]{@{}c@{}}Victim\\Model\end{tabular} 
        & Attack 
        & RNet
        & HNet
        & ENet
        & HNet$'$
        & ENet$'$
        & \begin{tabular}[c]{@{}c@{}}Target\\Extraction\end{tabular} 
        & PSNR$\uparrow$
        & MS-SSIM$\uparrow$
        & \begin{tabular}[r]{@{}r@{}}$\text{SR}_{\text{R}}\uparrow$\\or $\text{SR}_{\text{O}}\uparrow$\end{tabular}\\ \hline
  1  & \multirow{11}{*}{V1 \cite{wu2020watermarking}} & \multirow{3}{*}{EGG} & \multirow{3}{*}{UNet}  & \multirow{3}{*}{UNet} & \multirow{3}{*}{EEENet} & \multirow{3}{*}{--} & \multirow{3}{*}{--} & Blank & $41.81$ &$0.9992$ & $1.00$ \\ 
  2  & & & & & & & & ``PAMI''      & $33.94$ & $0.9920$ & $1.00$ \\ 
  3  & & & & & & & & ``COPYRIGHT'' & $32.97$ & $0.9908$ & $1.00$ \\ \cline{3-12}
  4  & & \multirow{8}{*}{Transferable} & \multirow{8}{*}{UNet} & \multirow{8}{*}{--} & \multirow{8}{*}{--} & CGAN & CEILNet & Blank & $28.73$ & $0.9836$ & $1.00$ \\ 
  5  & & & & & & CGAN     & EEENet  & Blank    & $31.38$ & $0.9870$  & $0.20$ \\ 
  6  & & & & & & Resnet50 & CEILNet & Blank    & $27.12$ & $0.9756$  & $1.00$ \\ 
  7  & & & & & & Resnet50 & EEENet  & Blank    & $30.11$ & $0.9796$  & $0.98$ \\ 
  8  & & & & & & UNet     & CEILNet & Blank    & $31.46$ & $0.9915$  & $1.00$ \\ 
  9  & & & & & & UNet     & EEENet  & Blank    & $41.09$ & $0.9987$  & $0.04$ \\ 
  10 & & & & & & CGAN     & CEILNet & Peppers  & $26.62$ & $0.9721$  & $0.00$ \\ 
  11 & & & & & & CGAN     & CEILNet & ``IEEE'' & $28.46$ & $0.9825$  & $0.00$ \\ \hline
  12 & \multirow{11}{*}{V2 \cite{zhang2021deep}} & \multirow{3}{*}{EGG} & \multirow{3}{*}{UNet}  & \multirow{3}{*}{UNet} & \multirow{3}{*}{CEILNet} & \multirow{3}{*}{--} & \multirow{3}{*}{--} & Blank & $34.81$ & $0.9939$ & $1.00$ \\ 
  13 & & & & & & & & ``PAMI''      & $32.35$ & $0.9894$ & $1.00$ \\ 
  14 & & & & & & & & ``COPYRIGHT'' & $33.67$ & $0.9916$ & $1.00$ \\ \cline{3-12}
  15 & & \multirow{8}{*}{Transferable} & \multirow{8}{*}{UNet} & \multirow{8}{*}{--} & \multirow{8}{*}{--} & CGAN & CEILNet & Blank & $30.62$ & $0.9874$ & $1.00$ \\ 
  16 & & & & & & CGAN     & EEENet  & Blank    & $32.49$ & $0.9889$ & $1.00$ \\ 
  17 & & & & & & Resnet50 & CEILNet & Blank    & $28.30$ & $0.9797$ & $1.00$ \\ 
  18 & & & & & & Resnet50 & EEENet  & Blank    & $32.12$ & $0.9830$ & $0.98$ \\
  19 & & & & & & UNet     & CEILNet & Blank    & $35.01$ & $0.9950$ & $1.00$ \\ 
  20 & & & & & & UNet     & EEENet  & Blank    & $10.05$ & $0.0155$ & $0.24$ \\
  21 & & & & & & CGAN     & CEILNet & Peppers  & $27.70$ & $0.9754$ & $0.00$ \\ 
  22 & & & & & & CGAN     & CEILNet & ``IEEE'' & $30.19$ & $0.9863$ & $0.00$ \\ \hline
  23 & \multirow{6}{*}{--} & -- & -- & \multirow{6}{*}{UNet} & EEENet & -- & -- & \multirow{6}{*}{Blank} & \multirow{6}{*}{--} & \multirow{6}{*}{--} & $1.00$ \\ 
  24 & & --           & --   & & CEILNet & --   & --      & & & & $1.00$ \\ 
  25 & & EGG          & UNet & & EEENet  & --   & --      & & & & $1.00$ \\ 
  26 & & EGG          & UNet & & CEILNet & --   & --      & & & & $1.00$ \\ 
  27 & & Transferable & UNet & & EEENet  & CGAN & CEILNet & & & & $1.00$ \\ 
  28 & & Transferable & UNet & & CeilNet & CGAN & CEILNet & & & & $1.00$ \\
  \hline
  \hline
  \end{tabular}}
  \end{table*}

\subsection{Victim Models against Previous Removal Attacks}
\label{sec:against_previous_removal_attacks}
We first assess the basic robustness of the victim models against previous black-box removal attacks including lattice attack presented in \cite{liu2023erase}, JPEG compression and noise addition, and the results are summarized in Fig. \ref{fig:lattice_attack}, Tables \ref{tab:JPEG_compression} and \ref{tab:Gaussian_noise}, respectively. Lattice attack replaces pixels in $b_i'$ at intervals of q with random values, thereby affecting ENet's ability to extract the watermark. We observe that the lattice attack can only successfully compromise V2 \cite{zhang2021deep} when q is lower than $11$, while it fails to compromise V1 \cite{wu2020watermarking} in all cases. Moreover, the image quality degradation when q is under $11$ is markedly perceivable even to the human eye, which is unacceptable for attacker.

For JPEG and noise addition, $b_i'$ is undergone JPEG compression or noise addition and fed into ENet, whose output is compared with the embedded watermark to obtain $\text{SR}_\text{E}$. Then, the PSNR and MS-SSIM are calculated between $b_i$ and $b_i'$, where $b_i'$ has been through either compression or noise addition. We observe that V1 \cite{wu2020watermarking} can to withstand moderate to low level distortions\footnote{Our implementation of V1 \cite{wu2020watermarking} did not consider using noisy images for robust training which was carried out in \cite{wu2020watermarking} as mentioned by the authors therein. Therefore, the actual robustness of V1 against image processing can be better than the results shown in Tables \ref{tab:JPEG_compression} and \ref{tab:Gaussian_noise}.}. However, surprisingly, we note that V2 \cite{zhang2021deep} is vulnerable against image processing. The underlying reason may lie in that it was developed to protect medical images which normally do not permit adding distortions.


\subsection{Quantitative Evaluation}
\label{sec:quantitative_eval}

\subsubsection{Quantitative Evaluation of the Proposed Attacks}
\label{sec:quantitaive evaluation of proposed attacks}
The quantitative evaluation results of the proposed attacks are summarized in Table \ref{tab:quantitative_result}. It contains a comprehensive analysis with $28$ experiments, among which experiments $1$--$11$ are against V1 \cite{wu2020watermarking}, and experiments $12$--$22$ are against V2 \cite{zhang2021deep}. Experiments $23$--$28$ are carried out to test false positives (or equivalently true negatives) in watermark extraction, i.e., given a non-watermarked input image, it tests that whether ENet or ENet$'$ will yield a ``Blank'' or a watermark image. We have the following observations. 1) Both the EGG and transferable removers can achieve high image quality (except experiment $20$), preventing visual detection that the watermarked images have been compromised. 2) While the EGG remover consistently delivers satisfactory attack results, it is not always the best performer. For instance, a comparison between Experiments $12$ and $19$ shows that to attack V2, the best transferable remover outperforms the best EGG remover in terms of image quality. 3) Regarding the architecture of the transferable remover $\text{ENet}'$, CEILNet is generally a better choice over EEENet. This is evidenced by comparing the $\text{SR}_{\text{R}}$ values between experiments $4$ and $5$, $19$ and $20$, etc. 4) V2 \cite{zhang2021deep} is more susceptible to attacks than V1 \cite{wu2020watermarking}, as indicated by the $\text{SR}_{\text{R}}$ comparisons between experiments $5$ and $16$, $9$ and $20$, etc., which is consistent with the qualitative results. 5) The victims models and the proposed removers are all reliable in rejecting false positives, which is evidenced from experiments $23$--$28$ that if queried by a non-watermarked image, the extraction output is a ``Blank'' image. Overall, results summarized in Table \ref{tab:quantitative_result} verify the effectiveness of the proposed attacks.

\subsubsection{Generalizability Test Against \cite{huang2023can}}
We now test the generalization capability of our proposed attacks under the situation that the victim model is completely unknown, indicating that the proxy model workflow is mismatched with the victim model workflow. We note that the work in \cite{huang2023can} does not have a hiding net and it only shares the same verification process. In other words, the framework in \cite{huang2023can} is different from our considered common box-free framework in Fig. \ref{fig:victim_model}. We are thus interested in testing whether an image generated by the protected generator in \cite{huang2023can}, after being processed by our proxy RNet$'$s, can still be verified with the correct ownership. The results are summarized in Table  \ref{tab:Generalizability results}, where the area under the curve (AUC) and average precision (AP) are used as the metrics. The results demonstrate that removers trained with any combination of HNet$'$ and ENet$'$ can significantly decrease the classifier's accuracy, while UNet+EEENet achieves the highest performance degradation. We believe the effectiveness of the remover stems from its ability to alter the position of samples in the decision space, thereby confusing the verification model's decision boundaries and leading to incorrect results.

\subsection{Ablation Study}
\label{sec:ablation_study}
We note that in (\ref{eq:concatenate}) and (\ref{eq:overwrite}), the input image is channel-wise concatenated with $\xi_0$ and $\xi_\text{Overwrite}$, respectively. While this is apparently necessary for watermark overwriting, it seems not to be a must if the objective is watermark removal as long as the supervision signal at the output is an all-white image, which represents no watermark, following the setting in \cite{zhang2021deep}. We thus demonstrate the necessity of the concatenation process in (\ref{eq:concatenate}) in this subsection. We consider V2 \cite{zhang2021deep} as an example, and the ablation study results for the EGG and transferable removers are summarized in Table \ref{tab:AS}, where ``--'' means no concatenation, and the transferable remover is CGAN plus CEILNet. The results reveal that concatenating $b_i'$ during both the training and testing phases yield the best results.

\begin{table}[!t]
\centering
\caption{Generalizability test results by launching transferable removers against V3 \cite{huang2023can}, where $0< \text{AUC} <1$, $0< \text{AP} <1$.}
\label{tab:Generalizability results}
\vspace*{-9pt}
\setlength{\tabcolsep}{4pt}
\begin{tabular}{c|c|cc|cc}
\hline
\hline
\multirow{2}{*}{HNet$'$} & \multirow{2}{*}{ENet$'$} & \multicolumn{2}{c|}{Before Attack} & \multicolumn{2}{c}{After Attack} \\
\cline{3-6}
& & AUC$\uparrow$ & AP$\uparrow$ & AUC$\uparrow$ & AP$\uparrow$ \\
\hline
\multirow{2}{*}{CGAN}     & CEILNet & \multirow{2}{*}{$0.99$} & \multirow{2}{*}{$0.99$} & $0.80$ & $0.75$  \\
                          & EEENet  &  &  & $0.82$ & $0.78$  \\ \hline
\multirow{2}{*}{Resnet50} & CEILNet & \multirow{2}{*}{$0.99$} & \multirow{2}{*}{$0.99$} & $0.94$ & $0.94$  \\
                          & EEENet  &  &  & $0.72$ &  $0.80$ \\ \hline
\multirow{2}{*}{UNet}     & CEILNet & \multirow{2}{*}{$0.99$} & \multirow{2}{*}{$0.99$} & $0.96$ & $0.95$  \\
                          & EEENet  &  &  & $0.60$ & $0.60$ \\
\hline
\hline
\end{tabular}
\end{table}

\begin{table}[!t]
  \centering
  \caption{Ablation study on image concatenation against V2 \cite{zhang2021deep}, where PSNR is in dB, $0< \text{MS-SSIM} <1$, and $0<\text{SR}<1$. If the ``Target Extraction'' is ``Blank'', then the last column uses $\text{SR}_\text{R}$ to test watermark removal. Otherwise, the last column uses $\text{SR}_\text{O}$ to test watermark overwriting.}
  \label{tab:AS} 
  \vspace*{-9pt}
  \setlength{\tabcolsep}{4pt}
  \resizebox{.99\columnwidth}{!}{
  \begin{tabular}{c|c|c|ccc}
  \hline
  \hline
  & \multicolumn{2}{c|}{Concatenated Image} 
  & \multirow{2}{*}{PSNR$\uparrow$} 
  & \multirow{2}{*}{MS-SSIM$\uparrow$} 
  & \multirow{2}{*}{\begin{tabular}[r]{@{}r@{}}$\text{SR}_{\text{R}}\uparrow$\\or $\text{SR}_{\text{O}}\uparrow$\end{tabular}} \\
  \cline{2-3}
  & Train & Test & & &\\ 
  \hline
  \multirow{4}{*}{\rotatebox[origin=c]{90}{EGG}} & Blank & Blank & $34.81$ & $0.9939$ & $1.00$ \\
  & Blank & ``COPYRIGHT'' & $24.76$ & $0.9320$ & $1.00$ \\
  & Blank & $a_{i}$       & $9.27 $ & $0.6535$ & $1.00$ \\
  & --    & --            & $33.24$ & $0.9917$ & $1.00$ \\
  \hline
  \multirow{4}{*}{\rotatebox[origin=c]{90}{{\footnotesize{Transferable}}}} & Blank & Blank & $30.62$ & $0.9974$ & $1.00$ \\
  & Blank & ``COPYRIGHT'' & $23.80$ & $0.9091$ & $0.96$ \\
  & Blank & $a_{i}$       & $14.55$ & $0.8883$ & $0.88$ \\
  & --    & --            & $30.58$ & $0.9868$ & $1.00$ \\
  \hline
  \hline
  \end{tabular}}
  \end{table}

\subsection{Limitations}
The primary limitations of our proposed removers are related to concerns about computational efficiency. For example, estimating a single gradient value of the EGG remover during training can require hundreds of queries. Additionally, optimizing the architecture of the transferable remover for peak performance necessitates continuous comparisons among several options. Despite numerous efforts to reduce query times for gradient estimation \cite{shi2022query, dong2021query} and accelerate neural architecture search \cite{ren2021comprehensive}, computational efficiency continues to be a significant challenge. Furthermore, this paper focuses solely on image processing tasks, which is a limited area within the broader scope of deep learning. Our future research will aim to improve the efficiency of our proposed attacks and expand the range of models considered.

\section{Conclusion}
We have uncovered the vulnerability of existing box-free model watermarks against black-box watermark removal attacks. Targeting the state-of-the-art box-free watermarking methods for low-level image processing DNNs, we have considered three black-box threat model scenarios, more practical and challenging than existing threat models, under which we have developed the EGG remover RNet under Scenarios 1 and 2, and the transferability-based remover RNet$'$ under the most challenging Scenario 3. Through extensive experiments against the victim models \cite{wu2020watermarking} and \cite{zhang2021deep}, we have rigorously demonstrated that the proposed removers can not only remove (both removers) but also overwrite (EGG remover) the watermarks. We have further demonstrated the generalization capability of the proposed transferable attacks under the workflow-mismatched situation, against the victim model in \cite{huang2023can}. Overall, our research findings indicate the urgent need for the development of removal-proof box-free watermarking methods in real-world situations.

\ifCLASSOPTIONcaptionsoff
  \newpage
\fi
\bibliographystyle{IEEEtran}

\bibliography{ref}

\begin{thebibliography}{10}
\providecommand{\url}[1]{#1}
\csname url@samestyle\endcsname
\providecommand{\newblock}{\relax}
\providecommand{\bibinfo}[2]{#2}
\providecommand{\BIBentrySTDinterwordspacing}{\spaceskip=0pt\relax}
\providecommand{\BIBentryALTinterwordstretchfactor}{4}
\providecommand{\BIBentryALTinterwordspacing}{\spaceskip=\fontdimen2\font plus
\BIBentryALTinterwordstretchfactor\fontdimen3\font minus \fontdimen4\font\relax}
\providecommand{\BIBforeignlanguage}[2]{{%
\expandafter\ifx\csname l@#1\endcsname\relax
\typeout{** WARNING: IEEEtran.bst: No hyphenation pattern has been}%
\typeout{** loaded for the language `#1'. Using the pattern for}%
\typeout{** the default language instead.}%
\else
\language=\csname l@#1\endcsname
\fi
#2}}
\providecommand{\BIBdecl}{\relax}
\BIBdecl

\bibitem{devlin2018bert}
J.~Devlin, M.-W. Chang, K.~Lee, and K.~Toutanova, ``Bert: Pre-training of deep bidirectional transformers for language understanding,'' \emph{arXiv preprint arXiv:1810.04805}, 2018.

\bibitem{dosovitskiy2020image}
A.~Dosovitskiy, L.~Beyer, A.~Kolesnikov, D.~Weissenborn, X.~Zhai, T.~Unterthiner, M.~Dehghani, M.~Minderer, G.~Heigold, S.~Gelly \emph{et~al.}, ``An image is worth 16x16 words: Transformers for image recognition at scale,'' \emph{arXiv preprint arXiv:2010.11929}, 2020.

\bibitem{lam2018automated}
C.~Lam, D.~Yi, M.~Guo, and T.~Lindsey, ``Automated detection of diabetic retinopathy using deep learning,'' \emph{AMIA summits on translational science proceedings}, vol. 2018, p. 147, 2018.

\bibitem{brown2020language}
T.~Brown, B.~Mann, N.~Ryder, M.~Subbiah, J.~D. Kaplan, P.~Dhariwal, A.~Neelakantan, P.~Shyam, G.~Sastry, A.~Askell \emph{et~al.}, ``Language models are few-shot learners,'' \emph{Advances in neural information processing systems}, vol.~33, pp. 1877--1901, 2020.

\bibitem{li2021survey}
Y.~Li, H.~Wang, and M.~Barni, ``A survey of deep neural network watermarking techniques,'' \emph{Neurocomputing}, vol. 461, pp. 171--193, 2021.

\bibitem{zhong2023brief}
X.~Zhong, A.~Das, F.~Alrasheedi, and A.~Tanvir, ``A brief, in-depth survey of deep learning-based image watermarking,'' \emph{Applied Sciences}, vol.~13, no.~21, p. 11852, 2023.

\bibitem{sharma2023review}
S.~Sharma, J.~J. Zou, G.~Fang, P.~Shukla, and W.~Cai, ``A review of image watermarking for identity protection and verification,'' \emph{Multimedia Tools and Applications}, pp. 1--63, 2023.

\bibitem{zhang2020model}
J.~Zhang, D.~Chen, J.~Liao, H.~Fang, W.~Zhang, W.~Zhou, H.~Cui, and N.~Yu, ``Model watermarking for image processing networks,'' in \emph{Proceedings of the AAAI conference on artificial intelligence}, vol.~34, no.~07, 2020, pp. 12\,805--12\,812.

\bibitem{wu2020watermarking}
H.~Wu, G.~Liu, Y.~Yao, and X.~Zhang, ``Watermarking neural networks with watermarked images,'' \emph{IEEE Transactions on Circuits and Systems for Video Technology}, vol.~31, no.~7, pp. 2591--2601, 2020.

\bibitem{zhang2021deep}
J.~Zhang, D.~Chen, J.~Liao, W.~Zhang, H.~Feng, G.~Hua, and N.~Yu, ``Deep model intellectual property protection via deep watermarking,'' \emph{IEEE Transactions on Pattern Analysis and Machine Intelligence}, vol.~44, no.~8, pp. 4005--4020, 2021.

\bibitem{huang2023can}
Z.~Huang, B.~Li, Y.~Cai, R.~Wang, S.~Guo, L.~Fang, J.~Chen, and L.~Wang, ``What can discriminator do? towards box-free ownership verification of generative adversarial networks,'' in \emph{Proceedings of the IEEE/CVF international conference on computer vision}, 2023, pp. 5009--5019.

\bibitem{chen2017zoo}
P.-Y. Chen, H.~Zhang, Y.~Sharma, J.~Yi, and C.-J. Hsieh, ``Zoo: Zeroth order optimization based black-box attacks to deep neural networks without training substitute models,'' in \emph{Proceedings of the 10th ACM workshop on artificial intelligence and security}, 2017, pp. 15--26.

\bibitem{dong2021query}
Y.~Dong, S.~Cheng, T.~Pang, H.~Su, and J.~Zhu, ``Query-efficient black-box adversarial attacks guided by a transfer-based prior,'' \emph{IEEE Transactions on Pattern Analysis and Machine Intelligence}, vol.~44, no.~12, pp. 9536--9548, 2021.

\bibitem{shi2022query}
Y.~Shi, Y.~Han, Q.~Hu, Y.~Yang, and Q.~Tian, ``Query-efficient black-box adversarial attack with customized iteration and sampling,'' \emph{IEEE Transactions on Pattern Analysis and Machine Intelligence}, vol.~45, no.~2, pp. 2226--2245, 2022.

\bibitem{lukas2022sok}
N.~Lukas, E.~Jiang, X.~Li, and F.~Kerschbaum, ``Sok: How robust is image classification deep neural network watermarking?'' in \emph{2022 IEEE Symposium on Security and Privacy (SP)}.\hskip 1em plus 0.5em minus 0.4em\relax IEEE, 2022, pp. 787--804.

\bibitem{guo2021fine}
S.~Guo, T.~Zhang, H.~Qiu, Y.~Zeng, T.~Xiang, and Y.~Liu, ``Fine-tuning is not enough: A simple yet effective watermark removal attack for dnn models,'' in \emph{Proc. IJCAI}, 2021.

\bibitem{uchida2017embedding}
Y.~Uchida, Y.~Nagai, S.~Sakazawa, and S.~Satoh, ``Embedding watermarks into deep neural networks,'' in \emph{Proceedings of the 2017 ACM on international conference on multimedia retrieval}, 2017, pp. 269--277.

\bibitem{darvish2019deepsigns}
B.~Darvish~Rouhani, H.~Chen, and F.~Koushanfar, ``Deepsigns: An end-to-end watermarking framework for ownership protection of deep neural networks,'' in \emph{Proceedings of the Twenty-Fourth International Conference on Architectural Support for Programming Languages and Operating Systems}, 2019, pp. 485--497.

\bibitem{li2021feature}
Y.~Li, L.~Abady, H.~Wang, and M.~Barni, ``A feature-map-based large-payload dnn watermarking algorithm,'' in \emph{International Workshop on Digital Watermarking}.\hskip 1em plus 0.5em minus 0.4em\relax Springer, 2021, pp. 135--148.

\bibitem{le2020adversarial}
E.~Le~Merrer, P.~Perez, and G.~Tr{\'e}dan, ``Adversarial frontier stitching for remote neural network watermarking,'' \emph{Neural Computing and Applications}, vol.~32, pp. 9233--9244, 2020.

\bibitem{fan2021deepipr}
L.~Fan, K.~W. Ng, C.~S. Chan, and Q.~Yang, ``Deepipr: Deep neural network ownership verification with passports,'' \emph{IEEE Transactions on Pattern Analysis and Machine Intelligence}, vol.~44, no.~10, pp. 6122--6139, 2021.

\bibitem{li2021spread}
Y.~Li, B.~Tondi, and M.~Barni, ``Spread-transform dither modulation watermarking of deep neural network,'' \emph{Journal of Information Security and Applications}, vol.~63, p. 103004, 2021.

\bibitem{hua23unambiguous}
G.~Hua, A.~B.~J. Teoh, Y.~Xiang, and H.~Jiang, ``Unambiguous and high-fidelity backdoor watermarking for deep neural networks,'' \emph{IEEE Transactions on Neural Networks and Learning Systems}, vol. Early Access, pp. 1--14, 2023.

\bibitem{adi2018turning}
Y.~Adi, C.~Baum, M.~Cisse, B.~Pinkas, and J.~Keshet, ``Turning your weakness into a strength: Watermarking deep neural networks by backdooring,'' in \emph{27th USENIX Security Symposium (USENIX Security 18)}, 2018, pp. 1615--1631.

\bibitem{zhang2018protecting}
J.~Zhang, Z.~Gu, J.~Jang, H.~Wu, M.~P. Stoecklin, H.~Huang, and I.~Molloy, ``Protecting intellectual property of deep neural networks with watermarking,'' in \emph{Proceedings of the 2018 on Asia conference on computer and communications security}, 2018, pp. 159--172.

\bibitem{namba2019exponential}
R.~Namba and J.~Sakuma, ``Robust watermarking of neural network with exponential weighting,'' in \emph{Proc. Asia Conference on Computer and Communications Security (ASIA CCS)}, 2019, pp. 228--240.

\bibitem{zhong2020protecting}
Q.~Zhong, L.~Y. Zhang, J.~Zhang, L.~Gao, and Y.~Xiang, ``Protecting ip of deep neural networks with watermarking: A new label helps,'' in \emph{Advances in Knowledge Discovery and Data Mining: 24th Pacific-Asia Conference, PAKDD 2020, Singapore, May 11--14, 2020, Proceedings, Part II 24}.\hskip 1em plus 0.5em minus 0.4em\relax Springer, 2020, pp. 462--474.

\bibitem{li2020protecting}
M.~Li, Q.~Zhong, L.~Y. Zhang, Y.~Du, J.~Zhang, and Y.~Xiang, ``Protecting the intellectual property of deep neural networks with watermarking: The frequency domain approach,'' in \emph{2020 IEEE 19th International Conference on Trust, Security and Privacy in Computing and Communications (TrustCom)}.\hskip 1em plus 0.5em minus 0.4em\relax IEEE, 2020, pp. 402--409.

\bibitem{quan2020watermarking}
Y.~Quan, H.~Teng, Y.~Chen, and H.~Ji, ``Watermarking deep neural networks in image processing,'' \emph{IEEE Transactions on Neural Networks and Learning Systems}, vol.~32, no.~5, pp. 1852--1865, 2020.

\bibitem{yu2021artificial}
N.~Yu, V.~Skripniuk, S.~Abdelnabi, and M.~Fritz, ``Artificial fingerprinting for generative models: Rooting deepfake attribution in training data,'' in \emph{Proceedings of the IEEE/CVF International conference on computer vision}, 2021, pp. 14\,448--14\,457.

\bibitem{yu2022responsible}
N.~Yu, V.~Skripniuk, D.~Chen, L.~Davis, and M.~Fritz, ``Responsible disclosure of generative models using scalable fingerprinting,'' in \emph{Proc. ICLR}, 2022.

\bibitem{Baluja2020hiding}
S.~Baluja, ``Hiding images within images,'' \emph{IEEE Transactions on Pattern Analysis and Machine Intelligence}, vol.~42, no.~7, pp. 1685--1697, 2020.

\bibitem{wang2022rethinking}
R.~Wang, H.~Li, L.~Mu, J.~Ren, S.~Guo, L.~Liu, L.~Fang, J.~Chen, and L.~Wang, ``Rethinking the vulnerability of dnn watermarking: Are watermarks robust against naturalness-aware perturbations?'' in \emph{Proceedings of the 30th ACM International Conference on Multimedia}, 2022, pp. 1808--1818.

\bibitem{liu2023erase}
H.~Liu, T.~Xiang, S.~Guo, H.~Li, T.~Zhang, and X.~Liao, ``Erase and repair: An efficient box-free removal attack on high-capacity deep hiding,'' \emph{IEEE Transactions on Information Forensics and Security}, vol.~18, pp. 5229--5242, 2023.

\bibitem{jiang2023evading}
Z.~Jiang, J.~Zhang, and N.~Z. Gong, ``Evading watermark based detection of ai-generated content,'' in \emph{Proceedings of the 2023 ACM SIGSAC Conference on Computer and Communications Security}, 2023, pp. 1168--1181.

\bibitem{you2024two}
J.~You and Y.~Zhou, ``Two-stage watermark removal framework for spread spectrum watermarking,'' \emph{IEEE Transactions on Multimedia}, vol.~26, pp. 7687--7699, 2024.

\bibitem{Niu2023fine}
L.~Niu, X.~Zhao, B.~Zhang, and L.~Zhang, ``Fine-grained visible watermark removal,'' in \emph{2023 IEEE/CVF International Conference on Computer Vision (ICCV)}, 2023, pp. 12\,724--12\,733.

\bibitem{Tian2024self}
C.~Tian, M.~Zheng, T.~Jiao, W.~Zuo, Y.~Zhang, and C.-W. Lin, ``A self-supervised cnn for image watermark removal,'' \emph{IEEE Transactions on Circuits and Systems for Video Technology}, pp. 1--1, 2024.

\bibitem{Tian2024Perceptive}
C.~Tian, M.~Zheng, B.~Li, Y.~Zhang, S.~Zhang, and D.~Zhang, ``Perceptive self-supervised learning network for noisy image watermark removal,'' \emph{IEEE Transactions on Circuits and Systems for Video Technology}, pp. 1--1, 2024.

\bibitem{johnson2016perceptual}
J.~Johnson, A.~Alahi, and L.~Fei-Fei, ``Perceptual losses for real-time style transfer and super-resolution,'' in \emph{Computer Vision--ECCV 2016: 14th European Conference, Amsterdam, The Netherlands, October 11-14, 2016, Proceedings, Part II 14}.\hskip 1em plus 0.5em minus 0.4em\relax Springer, 2016, pp. 694--711.

\bibitem{everingham2010pascal}
M.~Everingham, L.~Van~Gool, C.~K. Williams, J.~Winn, and A.~Zisserman, ``The pascal visual object classes (voc) challenge,'' \emph{International journal of computer vision}, vol.~88, pp. 303--338, 2010.

\bibitem{ronneberger2015u}
O.~Ronneberger, P.~Fischer, and T.~Brox, ``U-net: Convolutional networks for biomedical image segmentation,'' in \emph{Medical Image Computing and Computer-Assisted Intervention--MICCAI 2015: 18th International Conference, Munich, Germany, October 5-9, 2015, Proceedings, Part III 18}.\hskip 1em plus 0.5em minus 0.4em\relax Springer, 2015, pp. 234--241.

\bibitem{fan2017generic}
Q.~Fan, J.~Yang, G.~Hua, B.~Chen, and D.~Wipf, ``A generic deep architecture for single image reflection removal and image smoothing,'' in \emph{Proceedings of the IEEE International Conference on Computer Vision}, 2017, pp. 3238--3247.

\bibitem{isola2017image}
P.~Isola, J.-Y. Zhu, T.~Zhou, and A.~A. Efros, ``Image-to-image translation with conditional adversarial networks,'' in \emph{Proceedings of the IEEE conference on computer vision and pattern recognition}, 2017, pp. 1125--1134.

\bibitem{he2016deep}
K.~He, X.~Zhang, S.~Ren, and J.~Sun, ``Deep residual learning for image recognition,'' in \emph{Proceedings of the IEEE conference on computer vision and pattern recognition}, 2016, pp. 770--778.

\bibitem{wang2003multiscale}
Z.~Wang, E.~Simoncelli, and A.~Bovik, ``Multiscale structural similarity for image quality assessment,'' in \emph{The Thrity-Seventh Asilomar Conference on Signals, Systems {\&} Computers}, vol.~2, 2003, pp. 1398--1402.

\bibitem{ren2021comprehensive}
P.~Ren, Y.~Xiao, X.~Chang, P.-Y. Huang, Z.~Li, X.~Chen, and X.~Wang, ``A comprehensive survey of neural architecture search: Challenges and solutions,'' \emph{ACM Computing Surveys (CSUR)}, vol.~54, no.~4, pp. 1--34, 2021.

\end{thebibliography}
\vfill

\newpage
\begin{IEEEbiography}[{\includegraphics[width=1in,height=1.25in,clip,keepaspectratio]{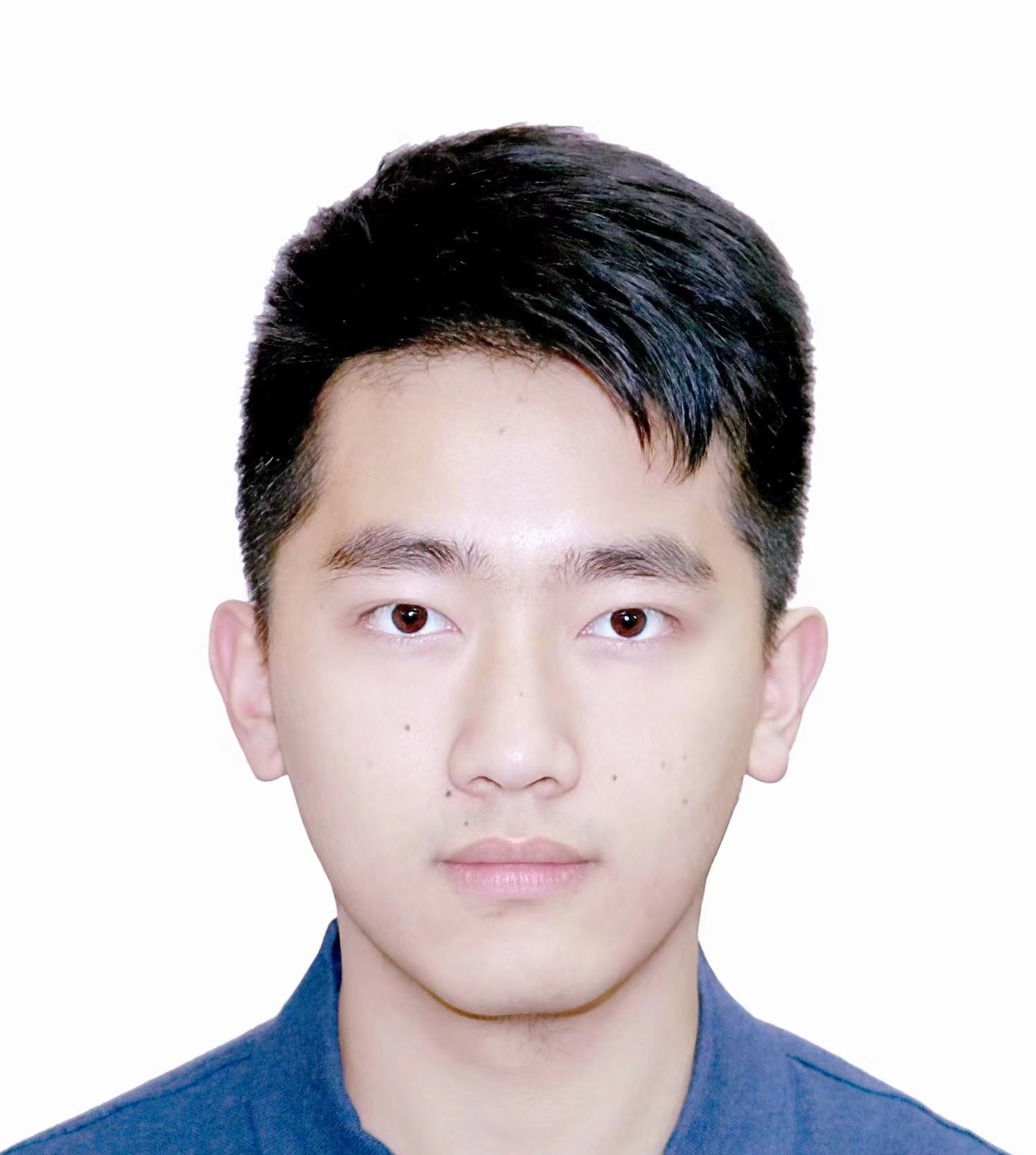}}]{Haonan An}
received his B.Eng. degree in Telecommunication Engineering from Huazhong University of Science and Technology, Wuhan, China, in 2023. He obtained his MS degree in Signal Processing from the School of Electrical and Electronic Engineering, Nanyang Technological University, Singapore, in 2024. He is currently pursuing his Ph.D. at the City University of Hong Kong. His research interests include AI security and autonomous driving.\end{IEEEbiography}

\begin{IEEEbiography}[{\includegraphics[width=1in,height=1.25in,clip,keepaspectratio]{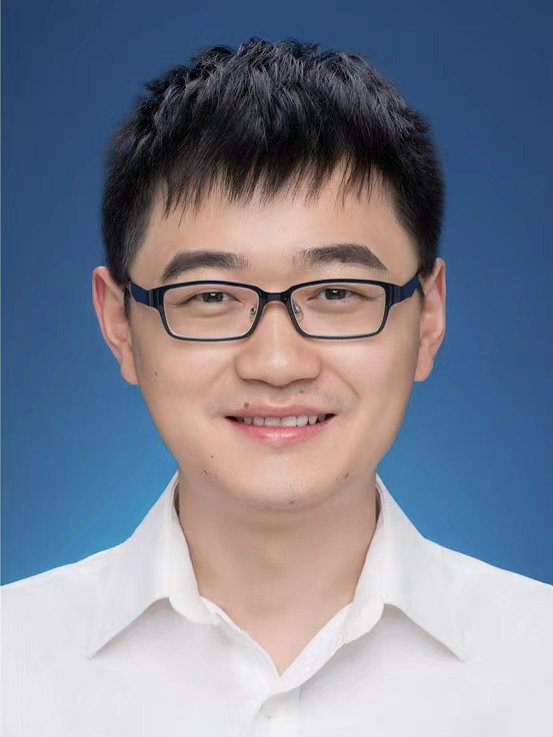}}]{Guang Hua} received the B.Eng. degree (Communication Engineering) from Wuhan University (WHU), China, in 2009, and the Ph.D. degree (Information Engineering) from Nanyang Technological University (NTU), Singapore, in 2014. He was a Research Fellow with the School of Electrical and Electronic Engineering, NTU (2015–2016), an Associate Professor with the School of Electronic Information, WHU (2017–2022), an International Scholar Exchange Fellow at Yonsei
University, sponsored by the CHEY Institute for Advanced Studies, South Korea (2020–2021), a Scientist (2013–2015) and
a Senior Scientist (2022–2023) with the Institute for Infocomm Research (I2R), A*STAR, Singapore. He is currently an Associate Professor with the Infocomm Technology (ICT) Cluster, Singapore Institute of Technology (SIT). His research interests include AI security, media forensics, applied machine learning, and general signal processing topics. He holds a Singapore patent (licensed) and a Chinese patent (transferred) on digital forensics. He serves as an Associate Editor for the IEEE Signal Processing Letters.\end{IEEEbiography}

\begin{IEEEbiography}[{\includegraphics[width=1in,height=1.25in,clip,keepaspectratio]{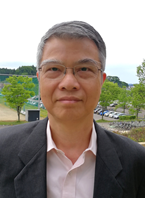}}]{Zhiping Lin} received B.Eng degree in control and automation from South China Institute of Technology, Guangzhou, China, Ph.D. degree in information engineering from the University of Cambridge, UK, in 1987. Since 1999, he has been an associate professor at the School of EEE, Nanyang Technological University (NTU), Singapore, where he is now serving as Programme Director of Master of Sciences Programme, the School of EEE, NTU. Prior to that, he worked at DSO National Labs, Singapore and Shantou University, China. Dr. Lin was the Editor-in-Chief of Multidimensional Systems and Signal Processing from 2011 to 2015, and a subject editor of the Journal of the Franklin Institute from 2015 to 2019. He also served as an associate editor for several other international journals, including IEEE Trans. on Circuits and Systems II. He was the coauthor of the 2007 Young Author Best Paper Award from the IEEE Signal Processing Society, and a Distinguished Lecturer of the IEEE Circuits and Systems Society during 2007-2008. His research interests include multidimensional systems, statistical signal processing and image/video processing, robotics and machine learning. He has published over 200 journal papers and over 200 conference papers.
\end{IEEEbiography}

\begin{IEEEbiography}[{\includegraphics[width=1in,height=1.25in,clip,keepaspectratio]{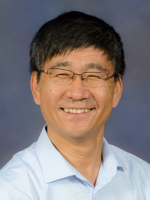}}]{Yuguang Fang}
received MS from Qufu Normal University, China, PhD from Case Western Reserve University, USA, and PhD from Boston University, USA, in 1987, 1994, and 1997, respectively. He joined the Department of Electrical and Computer Engineering at University of Florida in 2000 as an assistant professor, then was promoted to associate professor, full professor, and distinguished professor, in 2003, 2005, and 2019, respectively. Since 2022, he has been a Global STEM Scholar and Chair Professor with Department of Computer Science, City University of Hong Kong. 

He received many awards including US NSF CAREER Award, US ONR Young Investigator Award, 2018 IEEE Vehicular Technology Outstanding Service Award, and IEEE Communications Society awards (AHSN Technical Achievement Award, CISTC Technical Recognition Award, and WTC Recognition Award). He was Editor-in-Chief of IEEE Transactions on Vehicular Technology and IEEE Wireless Communications. He is a fellow of ACM, IEEE, and AAAS.  
\end{IEEEbiography}
\end{document}